\documentclass[twocolumn]{article}

\usepackage{framed,multirow}
\usepackage{amssymb}
\usepackage{latexsym}
\usepackage{soul}
\usepackage{booktabs}
\usepackage[markup=underlined]{changes}
\usepackage{url}
\usepackage{xcolor}
\usepackage{comment}
\usepackage{graphicx}
\usepackage{hyperref}
\usepackage[switch,pagewise]{lineno}
\usepackage{amsmath,amssymb,amsfonts}
\usepackage[numbers]{natbib}

\definecolor{newcolor}{rgb}{.8,.349,.1}

\usepackage{mathrsfs}
\usepackage[scr=boondoxo]{mathalfa}

\renewcommand{\L}{\mathcal{L}}

\newcommand{\f}{f}

\newcommand{\R}{\mathbb{R}}
\newcommand{\grad}[1]{\nabla#1}

\newcommand{\gt}[1]{\mathscr{#1}}

\definechangesauthor[color=red]{Rev1}
\definechangesauthor[color=teal]{Rev2}
\definechangesauthor[color=blue]{Rev3}
\definechangesauthor[color=violet]{Rev4}

\usepackage{todonotes}

\renewcommand{\dot}[2]{\big\langle#1,#2\big\rangle}
\newcommand\norm[1]{\left\lVert#1\right\rVert}
\newcommand{\lossPDE}{\mathcal{L}_{\text{LSE}}}
\newcommand{\lossdata}{\mathcal{L}_{\text{data}}}

\renewcommand{\dot}[2]{\left\langle#1,#2\right\rangle}

\newcommand{\hess}[1]{\mbox{\textbf{H}} #1}

\title{Geometric implicit neural representations for signed distance functions}%
\author{
  Luiz Schirmer\thanks{Corresponding author: \texttt{luiz.schirmer@ufsm.br}}
  \and Tiago Novello
  \and Vinícius da Silva
  \and Guilherme Schardong
  \and Daniel Perazzo
  \and Hélio Lopes
  \and Nuno Gon\c{c}alves
  \and Luiz Velho
}

\begin{document}



\maketitle











\begin{abstract}
\textit{Implicit neural representations} (INRs) have emerged as a promising framework for representing signals in low-dimensional spaces. This survey reviews the existing literature on the specialized INR problem of approximating \textit{signed distance functions} (SDFs) for surface scenes, using either oriented point clouds or a set of posed images. We refer to neural SDFs that incorporate differential geometry tools, such as normals and curvatures, in their loss functions as \textit{geometric} INRs. The key idea behind this 3D reconstruction approach is to include additional \textit{regularization} terms in the loss function, ensuring that the INR satisfies certain global properties that the function should hold -- such as having unit gradient in the case of SDFs.
We explore key methodological components, including the definition of INR, the construction of geometric loss functions, and sampling schemes from a differential geometry perspective. Our review highlights the significant advancements enabled by geometric INRs in surface reconstruction from oriented point clouds and posed images.


\end{abstract}

\textbf{Keywords:} Neural Fields, Implicit Representations, Differential Geometry


\section{Introduction}
\label{sec:intro}
An \textit{implicit neural representation} (INR) is a \textit{neural network} that parameterizes a signal in a low-dimensional domain. This representation differs from classical methods, as it encodes the signal implicitly in its parameters by mapping coordinates to target signal values. For example, in the case of an implicit surface, an INR $f$ takes a 3D point $p$ and returns the isosurface value $f(p)$. In this scenario, we aim for the INR to approximate the input data as closely as possible, similar to the problem of approximating signals using \textit{radial basis functions}. INRs provide a compact, high-quality, and smooth approximation for discrete data. Furthermore, INRs allow calculating higher-order derivatives in closed form through automatic differentiation, which is present in modern machine learning~frameworks.

INRs are smooth, compact networks that are fast to evaluate and have a high representational capacity. This has motivated their use in several contexts for example: images \cite{chen2021learning, paz2022}, face morphing~\cite{schardong2024neural, zheng2022imface, zheng2023imfacepp}, signed/unsigned distance functions~\cite{park2019deepsdf, novello2022exploring, Schirmer2023how, lindell2022bacon, sitzmann2020implicit, yang2021geometry, schirmer2021neural, silva2022mip, fainstein2024dudf, sang2023enhancing}, displacement fields~\cite{yifan2021geometry}, surface animation~\cite{mehta2022level, novello2023neural}, multiresolution signals \cite{paz2023mr, saragadam2022miner, lindell2022bacon}, occupancy \cite{mescheder2019occupancy}, constructive solid geometry~\cite{marschner2023constructive}, radiance fields~\cite{mildenhall2020nerf, perazzo2023directvoxgo++}, textures~\cite{paz2024implicit}, 3D reconstruction from images and videos~\cite{wang2021neus, wang2023neus2, li2023neuralangelo}, among others. These works leverage the fact that INRs are compositions of smooth maps to explore their derivatives during training.
From these various applications, geometry processing is a noteworthy field with many applications~\cite{wang2021neus, novello2022exploring, lindell2022bacon, wang2023neus2}.

The parameters \(\theta\) of an INR \(f\) are \textit{implicitly} defined as the solution to a non-linear equation \(\mathcal{L}(\theta) = 0\), where \(\mathcal{L}\) is a \textit{loss function} that ensures \(f\) fits the samples \(\{p_i, \gt{f}(p_i)\}\) of the ground-truth function \(\gt{f}\) and satisfies certain properties held by \(\gt{f}\). For instance, when fitting \(f\) to the \textit{signed distance function} (SDF) of a surface, a term is added to the loss function to enforce the gradient \(\nabla f\) of the network to be unitary; the Eikonal equation \(\|\nabla f\| = 1\). This is a fundamental concept in INRs, as SDFs are solutions to this partial differential equation. The primary benefit of adding this constraint is that the sampling \(\{p_i, \gt{f}(p_i)\}\) is often sparse and concentrated near the ground-truth surface. Consequently, training \(f\) only on these samples could introduce noise in regions with no data. Imposing the Eikonal equation on additional points helps regularizing the INR training.

In a neural SDF \(f\), the output \(f(p)\) is a distance value that can be positive or negative, indicating whether a point is inside or outside the underlying compact surface. A distance value of zero indicates that the point lies on the implicit surface \(S\). The gradient \(N = \nabla f\) provides the \textit{normal} field of \(S\), and its Hessian \(\mathrm{Hess}(f)\), the \textit{shape operator}, gives the curvatures.
In this work, we present a survey on \textit{geometric} approaches that explore these differential objects during the training and inference of INRs.

We define a \textit{geometric} INR as a neural network \(f: \mathbb{R}^3 \to \mathbb{R}\) approximating a SDF of a regular surface $S$, i.e. \(\|\nabla f\| \approx 1\), such that its parameters \(\theta\) are implicitly defined by \(\mathcal{L}(\theta) = 0\), with \(\mathcal{L}\) enforcing geometrical properties~of~\(S\) through $\nabla f$ and $\hess{f}$.
To enforce the SDF property, an Eikonal term $\int_{\Omega} (\|\nabla f\|-1)^2 dp$ is added to $\mathcal{L}$, where $\Omega$ is the training domain.
Another important geometric term arises from forcing the alignment of the normals $N$ of $S$ with the gradient $\nabla f$, i.e. $\int_{S} \big(1-\dot{\nabla f}{N}\big) dS$.

To bring an in-depth discussion about geometric INRs we consider the following \textit{training pipeline}.
It begins with the \textbf{input data}, which could be either an oriented point cloud consisting of points and normals sampled from the underlying surface $S$, or a set of posed images taken from a scene having $S$ as a surface.
Next, a \textbf{neural network} (INR) $f:\mathbb{R}^3\to \mathbb{R}$ with parameters $\theta$ is defined to fit the SDF of~$S$. This fitting is achieved through optimization of a \textbf{geometric loss function} $\mathcal{L}$ using a variant of the gradient descent algorithm.
However, computing the gradient $\nabla \mathcal{L}$ may be infeasible in practice due to the size of the data set. Thus, it is common to consider mini-batches (\textbf{sampling} step) which exploit geometric properties of the underlying surface $S$ to speed up the training. Once the INR $f$ is trained, the SDF properties of $f$ can be leveraged for various applications, such as geometry \textbf{inference} using \textit{sphere tracing} or surface evolution using \textit{level-set methods}.



We present recent frameworks that enhance the training performance of INRs by exploring geometrical losses and curvature information to sample points during training. Additionally, we discuss approaches that utilize geometric INRs for 3D reconstruction from posed images, where the neural SDF is used to represent the scene geometry. Finally, we provide examples of dynamic geometric INR approaches for learning surface animation from oriented point clouds and (time-dependent) posed images. To achieve this, the network domain must be extended to space-time \(\mathbb{R}^3 \times \mathbb{R}\) to encode the time variable \cite{novello2023neural,mehta2022level}.


\paragraph{\textbf{Terminology}}

In the visual computing community, implicit neural representations have also been referred to as neural fields, neural implicits, and coordinate-based neural networks. In this paper, we focus on the terminology ``implicit neural representations" despite some references using the other terms.

\vspace{0.2cm}

The paper is organized as follows. Section \ref{s-inr} discusses the main aspects of implicit surface reconstruction, focusing on the application of the Eikonal equation, oriented point-cloud-based reconstruction and classical image-based approaches. Section \ref{s-framework} shows a geometric framework to solve the geometric implicit neural representation problem, where we detail the input data, loss function details, and dataset sampling. Section \ref{s-implicits} presents applications considered state-of-the-art for INRs, where we focus on neural implicit surface reconstruction from images, from oriented point clouds, multi-resolution, and dynamic INRs with applications for deforming objects and animation. 
The final remarks are drawn at the conclusion in Section \ref{s-conclusion}.

\section{Implicit surface reconstruction}
\label{s-inr}
 Implicit representations are commonly used in computer graphics to represent 3D shapes. Unlike explicit representations (e.g. using triangle meshes) implicit representations encode a surface $S$ as the zero-level set of a function $\gt{f}:\mathbb{R}^3 \rightarrow \mathbb{R}$. For the surface $S$ to be regular, the zero must be a regular value of $\gt{f}$, that is, $\nabla \gt{f}\neq 0$ on $S=\gt{f}^{-1}(0)$. Again, SDFs are a common example of an implicit representation, where $\gt{f}$ is the solution of the Eikonal equation:
\begin{align}\label{e-eikonal}
    \|\nabla \gt{f}\|=1 \text{ subject to } \gt{f}=0 \text{ on } S.
\end{align}

In this work, we present recent strategies to solve \eqref{e-eikonal} by parameterizing $\gt{f}$ with an INR $f:\mathbb{R}^3\to \mathbb{R}$, with parameters~$\theta$. To approximate a solution of this equation, it is common to define a loss function $\mathcal{L}$ to enforce $f$ to be a solution. Solving this equation reveals that $\dot{\nabla \gt{f}}{N}=1$ on $S$, indicating that $\nabla \gt{f}$ aligns with the normals $N$ of $S$. We refer to a solution of the above problem as a \textit{geometric~INR}.

Before presenting examples of training pipelines for geometric INRs, we recall some classic approaches.

\subsection{Oriented point cloud based reconstruction}
\textit{Radial basis functions} (RBFs) \cite{carr2001reconstruction} is a classical method that can be used to approximate the SDF of a surface $S$ from a sample $\{p_i, \gt{f}_i\}$ of this function. The RBF is expressed as $s(p)=\sum\lambda_i\phi(\|p-p_i\|),$ where the coefficients $\lambda_i\in\mathbb{R}$ are determined by imposing $s(p_i)=\gt{f}_i$. The \textit{radial function} $\phi:\mathbb{R}^+\to \mathbb{R}$ is a real function and $p_i$ are the centers of the RBF~\cite{novello2022exploring}.
Note that the RBF representation depends on the data since its interpolant $s$ depends on the input points $p_i$.

\textit{Poisson surface reconstruction}~\cite{kazhdan2006poisson} is another classical method widely used in computer graphics to reconstruct a surface from an oriented point cloud~$\{p_i,N_i\}$. It revolves around solving the \textit{Poisson's equation}, using $\{p_i,N_i\}$.
The objective is to reconstruct an implicit function $f$ of the underlying surface by asking it to be zero at $p_i$ and to have gradients at $p_i$ equal to $N_i$. The pairs $\{p_i, N_i\}$ are used to define a vector field $V$.
Then, $f$ is computed by optimizing $\min_f \norm{\nabla f-V}$ which results in a Poisson problem: $\Delta f= \nabla\cdot V$.

\subsection{Image-based reconstruction}
There are many classical works that aim to reconstruct the surface \(S\) of a 3D scene from a set of unordered images \(\{I_j\}\) \cite{szeliski2022computer}. Generally, these methods focus on obtaining an oriented point cloud \(\{p_i, N_i\}\) using \textit{structure-from-motion} \cite{ullman1979interpretation}. The surface \(S\) can then be reconstructed using Poisson surface reconstruction \cite{kazhdan2006poisson}. COLMAP \cite{schoenberger2016sfm} is a standard example of this approach. It extracts features from each \(I_j\), e.g., using SIFT \cite{lowe1999object}, and searches for feature correspondences between the images, using RANSAC~\cite{fischler1981random}. Finally, using bundle adjustment, it computes camera positions and the points \(p_i\) such that the corresponding viewing rays intersect.

Recently, implicit neural representations initially developed for the novel view synthesis problem such as NeRFs~\cite{mildenhall2020nerf} have been gaining popularity for representing these systems.
Recently, adaptations for NeRFs have been created for the task of implicit surface reconstruction~\cite{wang2021neus,wang2023neus2, yariv2021volume, long2022sparseneus}.
We will present this problem in more detail in Subsection~\ref{sec:nerf_based}

\section{Geometric INR framework}
\label{s-framework}

\subsection{Overview of the problem}
%

\begin{figure}[h!]
    \centering
        \includegraphics[width=1\columnwidth]{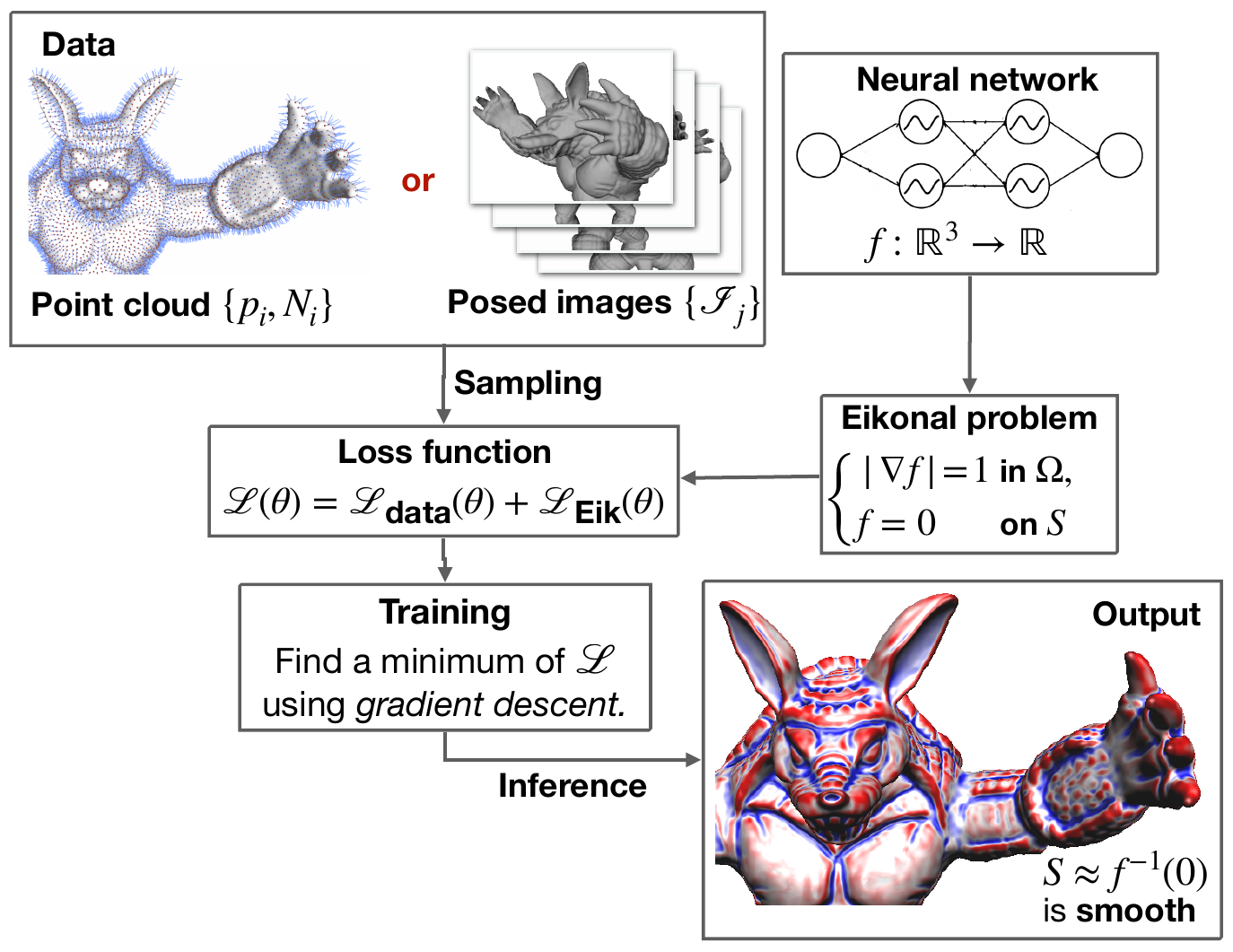}
        \caption{
Geometric INR pipeline: The input data can be either an oriented point cloud $\{p_i, N_i\}$ or a set of posed images $\{\gt{I}_j\}$. A neural network \(f\) is then defined to fit a solution to the Eikonal equation. To train \(f\), we define a loss function consisting of two terms: data constraint and Eikonal constraint. For the point-based data, we simply enforce \(f(p_i) = 0\) and \(\nabla f(p_i) = N_i\). For the image-based data, we rely on volume rendering techniques. Finally, gradient descent is used to optimize the resulting loss function.
        }
    \label{f-pipeline}
\end{figure}

This section presents an overview of the framework used to solve the geometric INR problem of training the parameters $\theta$ of an INR $f:\R^3\to \R$ to approximate the SDF of a desired surface $S$. This pipeline describes the problem for both point-based and image-based surface reconstruction. To present this pipeline we follow the scheme in Figure~\ref{f-pipeline}.

The \textbf{input} can be either a sample of oriented points $\{p_i, N_i\}_{i=1}^n$ from the \textit{ground truth} surface $S$ or a set of posed images $\{\gt{I}_j\}$ taken from a scene having $S$ as its surface. The \textbf{output} is an INR $f$ approximating the SDF of~$S$.
To estimate an SDF with its zero-level set of $S$ from $\gt{I}_j$, it is assumed that the existence of a projection invertible matrix transformation $W_j$ for each image, mapping from screen to world coordinates.
Although the problem seems fundamentally different, we shall discuss their similarities later.
The framework explores the normals $N$ to define a loss function and the curvatures of the surface to sample the mini-batches.


Next, we define a \textbf{neural network} (INR) $f$, with parameters~$\theta$, to fit the SDF $\gt{f}$ of the ground-truth surface $S$. For this, we define a \textbf{loss function} $\mathcal{L}=\mathcal{L}_{\text{data}}+ \mathcal{L}_{\text{Eik}}$ to enforce $f$ to be a solution of Eikonal equation~\eqref{e-eikonal}. The data term $\mathcal{L}_{\text{data}}$ forces $f$ to fit the input data. $\mathcal{L}_{\text{Eik}}$ forces $f$ to be a solution of the Eikonal equation; thus, it works like a (implicit) regularization.

For image-based reconstruction, the data term \(\mathcal{L}_{\text{data}}\) is modeled using volume rendering \cite{wang2021neus, wang2023neus2}. This involves defining a differentiable rendering function \(I(\theta, W_j)\) for the camera corresponding to each posed image \((\gt{I}_j, W_j)\). Thus, the data constraint \(\mathcal{L}_{\text{data}}\) for training \(\theta\) is defined by forcing \(I(\theta, W_j) = \gt{I}_j\). Some techniques also incorporate geometric regularizations, such as curvature-based methods \cite{li2023neuralangelo}, or different approaches to map the SDF to a density-based operator \cite{yariv2021volume}.

The \textbf{training} step consists of using a variant of the gradient descent algorithm to find a minimum of $\mathcal{L}$. However, in practice, computing the gradient $\nabla \mathcal{L}(\theta)$ may be unfeasible; thus, we consider \textbf{sampling} minibatches of the input data. Once we have the INR $f$ trained, we can infer its geometry to \textbf{render} its zero-level set $f^{-1}(0)$.


The following sections present each component of geometric INR training in detail.

\subsection{Input data}

Here, we describe the two kinds of data options (input of the pipeline in Figure~\ref{f-pipeline}) to reconstruct the surface $S$.

\paragraph{Oriented point cloud}
Let \(\{p_i, N_i\}_{i=1}^n\) be an oriented point cloud sampled from \(S\), where \(p_i \in S\) and \(N_i\) are the normals to \(S\) at~\(p_i\). We aim to reconstruct the SDF \(\gt{f}\) of \(S\) by enforcing \(f(p_i) = 0\) and \(\nabla f(p_i) = N_i\). However, this approach may result in a neural SDF with spurious components on its zero-level set. To mitigate such noise, we sample additional points \(\{q_k\}\) outside \(\{p_i\}\) to regularize the training. As a result, we obtain a set of points being the union of $\{p_i\}$ and $\{q_k\}$ with their corresponding SDF values.


\paragraph{Posed images}
For the image-based setting, we assume a set of images \(\gt{I}_j\) with their corresponding projection matrices \(W_j\). COLMAP~\cite{schoenberger2016sfm} is commonly used whenever the projection matrices are unavailable. \(W_j\) is the product \(K_j \cdot [R_j \,|\, t_j]\) of the intrinsic matrix \(K_j \in \mathbb{R}^{4 \times 4}\) and the extrinsic matrix \([R_j \,|\, t_j]\).
The intrinsic matrix \(K_j\) includes parameters of the camera, such as focal length and center of projection. The extrinsic matrix \([R_j \,|\, t_j]\) consists of the orthogonal matrix \(R_j\) for camera orientation and the camera position \(-t_j\). Note that in COLMAP, the images are assumed to have significant overlap, otherwise the feature matching step fails, resulting in poor approximations of \(W_j\).

\subsection{Network architecture}


We assume the INR $f:\R^3\to \R$ to be parametrized by a \textit{multilayer perceptron} (MLP) defined as follows.
\begin{equation}\label{e-network-architecture}
    \f(p)=W_n\circ f_{n-1}\circ f_{n-2}\circ \cdots \circ f_{0}(p)+b_n
\end{equation}
\noindent
where $f_{i}(p_i)=\varphi(W_i p_i + b_i)$ is the $i$th layer, and $p_i$ is the output of $f_{i-1}$, i.e. $p_i=f_{i-1}\circ \cdots \circ f_{0}(p)$. Here we apply the smooth activation function $\varphi:\R\to \R$ to each coordinate of the affine map, which is formed by the linear map $W_i:\R^{N_i}\to \R^{N_{i+1}}$ and the bias $b_i\in\R^{N_{i+1}}$. The operators $W_i$ are represented as matrices, and $b_i$ as vectors, combining their coefficients to form the parameters $\theta$ of the function $f$.
In the following section we define a loss function to train $\theta$ to fit $f$ to the input data.

The choice of the activation function $\varphi$ has a great impact on the representation capacity of $f$. For example, using sines, that is $\varphi=\sin$, results in a powerful INR architecture for surface reconstruction from oriented point clouds~\cite{sitzmann2020implicit,novello2022exploring}.

For image-based reconstruction, it is common to use Fourier feature mapping \cite{tancik2020fourier} to represent the neural SDF~\cite{wang2021neus}. In this approach, the activation function is \(\varphi = \texttt{ReLU}\), and the first layer projects the input onto a list of sines and cosines. However, this method often results in slow training time.
To speed up training and rendering times, hashgrid-based representation was proposed in \cite{muller2022instant, li2023neuralangelo, wang2023neus2}.

\subsection{Loss function}
\label{sec:loss}

We now define a loss function $\L$ to train the parameters $\theta$ of the INR $f$. Again, we start with the oriented point-based case.

\paragraph{Oriented point-based rendering}
We use the input data $\{p_i, N_i\}_{i=1}^n$ sampled from a surface $S$ and the Eikonal equation~\eqref{e-eikonal} to define the loss function $\L$
as the composition of $\L_{\text{data}}$, and $\L_{\text{Eik}}$. This loss is used to optimize $\theta$ such that $\f:\R^3\to\R$ approximates the SDF $\gt{f}$ of $S$.
\begin{align}\label{e-loss_function}
    \mathcal{L}(\theta)\!=\!
     \underbrace{\frac{1}{n}\sum_{i}f(p_i)^2 + \Big(1\!-\!\dot{\nabla f(p_i)}{N_i}\Big)}_{\mathcal{L}_{\text{data}}}
     \!+\!\!
     \underbrace{\int\limits_{\Omega}\!\!\!\big(1\!-\!\norm{\grad{f}}\big)^2dp}_{\mathcal{L}_{\text{Eik}}}.
\end{align}
Here, ${\mathcal{L}_{\text{Eik}}}$ encourages $\f$ to be the SDF of a set $\mathcal{X}$ by ensuring that $\norm{\grad{\f}}\!=\!1$, ${\mathcal{L}_{\text{data}}}$ encourages $\mathcal{X}$ to contain $S$; i.e. $f\!=\!\gt{f}$~on~$\{p_i\}$. In addition, it asks for the alignment between $\grad{\f}$ and the normals of $S$ to regularize the orientation near the zero-level set.

Typically, an additional term is added to \eqref{e-loss_function} to penalize points outside $S$, forcing $\f$ to be the SDF of $S$ (i.e., $\mathcal{X}=S$)~\cite{sitzmann2020implicit}.
A common approach is to extend ${\mathcal{L}_{\text{data}}}$ to consider (off-surface) points outside $S$ by using an approximation of the SDF of $S$~\cite{novello2022exploring}.
The SDF approximation at a point $p$ can be computed using $|\gt{f}(p)| \approx \min_{i} \norm{p - p_i}$.
The sign of $\gt{f}(p)$ at a specific point $p$ is negative if $p$ lies inside the surface $S$ and positive otherwise.
Note that for each point $p_i$ with a normal $N_i$, the sign of $\dot{p-p_i}{N_i}$ indicates the side of the tangent plane that $p$ belongs to \cite{novello2022exploring}. Therefore, we can estimate the sign of $\gt{f}(p)$ by adopting the dominant signs of the numbers $\dot{p-p_j}{N_j}$, with $\{p_j\}$ being a subset of the point clound.

\paragraph{Image-based rendering}
The traditional image-based pipeline for surface reconstruction converts the SDF $f$ to a volume density function $\sigma$ by composing $f$ with a density distribution function $\rho$, that is, $\sigma=\rho\circ f$.
Then, volume rendering is applied to create an image given a camera view.
Getting the intensity for a pixel $p$ along a view direction $v$ can be computed using \textit{ray marching}: where we march along a ray $r(t)=o+tv$ and accumulate the colors $c\big(r(t)\big)$ and densities $\sigma\big(r(t)\big)$~\cite{kajiya1984ray}.
The integral is done in the \textit{near} and \textit{far ray} bounds: $t_n$ and $t_f$.
The volume rendering equation is given by:
\begin{equation}\label{e-rendering_equation}
   I(r) = \int_{t_n}^{t_f} c\big(r(t)\big)\sigma\big(r(t)\big)\exp{\left(\int_{t_n}^{t} \sigma\big(r(s)\big)ds\right)} dt.
\end{equation}
To solve \eqref{e-rendering_equation}, most methods discretize the ray and employ numerical methods~\cite{max1995optical}. This results in a differentiable rendering which allows us to optimize the model parameters $\theta$~\cite{mildenhall2020nerf}.

After rendering each pixel corresponding to a input image $\gt{I}_j$ with projection matrix $W_j$, we can compare the output image $I(\theta,W_j)$ with its ground-truth $\gt{I}_j$ resulting in a \textit{Photometric loss}:
\begin{equation}\label{e-image-loss-data}
   {\mathcal{L}_{\text{data}}}(\theta) = \frac{1}{N}\sum_j^m\norm{I(\theta, W_j) - \gt{I}_j}^2.
\end{equation}

\subsection{Sampling}\label{s-sampling}

\paragraph{Point-based sampling}
Let \(\{p_i, N_i\}_{i=1}^n\) be a sample of the ground-truth surface \(S\). During optimization, we may not be able to compute the gradient \(\nabla_\theta \mathcal{L}_{\text{data}}(\theta)\) considering the whole dataset. Therefore, it is common to divide it into minibatches. \citet{novello2022exploring} considered the principal curvature information to prioritize regions with more geometric variations of the data during minibatch selection.
Regions of higher absolute principal curvatures encode more detailed information than regions with lower absolute curvatures.

\begin{figure}[h!]
    \centering
        \includegraphics[width=1\columnwidth]{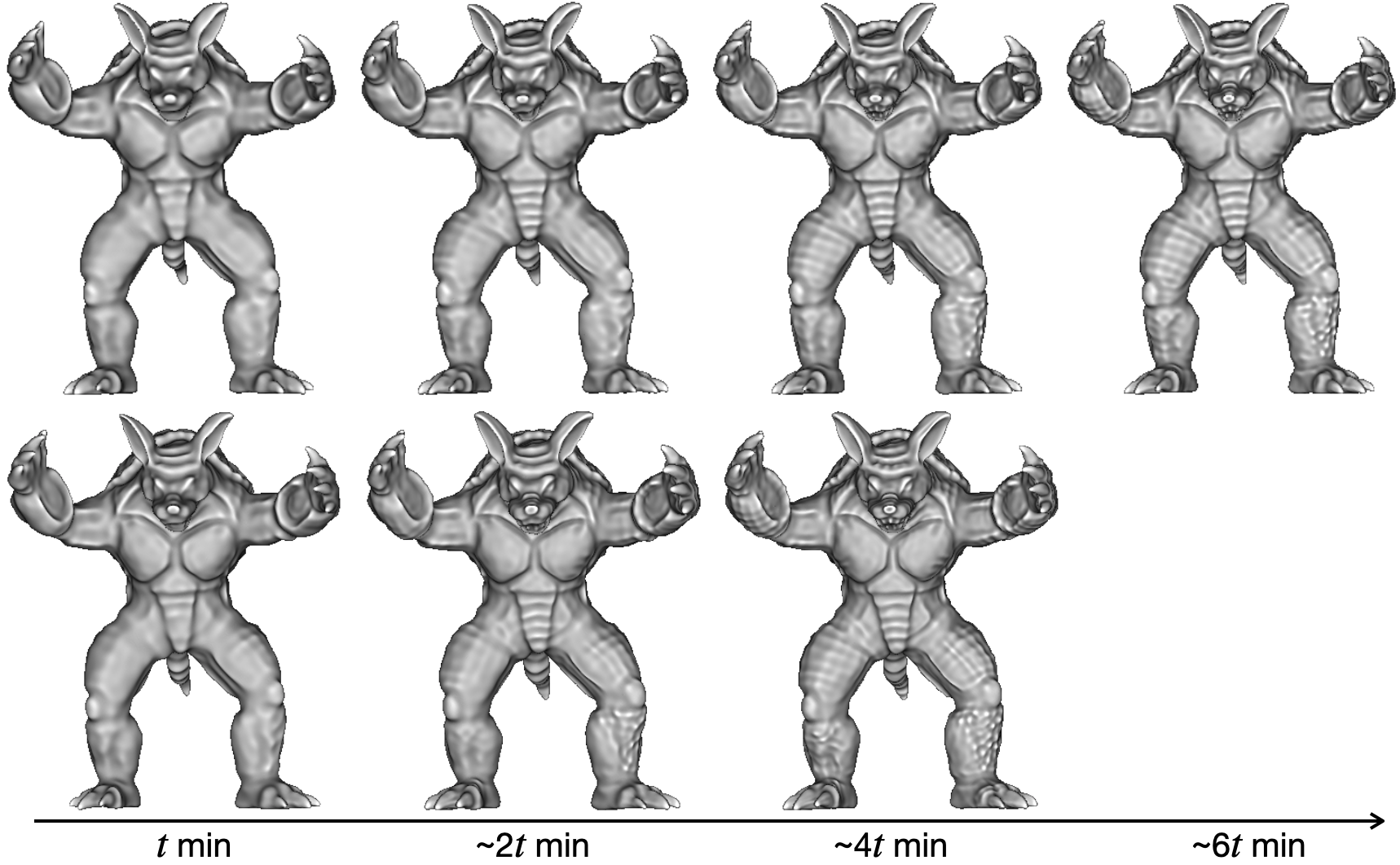}
        \caption{Neural implicit surfaces approximating the Armadillo model. The columns indicate the zero-level sets after $29$, $52$, $76$, and $100$ epochs of training.
        Line $1$ shows the results using minibatches sampled uniformly.
        Line $2$ presents the results using the adapted sampling of minibatches with $10\%$ / $70\%$ / $20\%$ of points with low/medium/high~features. Image from \cite{novello2022exploring}.
        }
    \label{f-biased_sampling}
\end{figure}

\citet{novello2022exploring} proposed splitting \(\{p_i\}\) into three sets based on their curvatures: \(V_1\) (low), \(V_2\) (medium), and \(V_3\) (high). During training, they prioritized the points with more geometrical information in \(V_2\) and \(V_3\) while sampling fewer points from \(V_1\) to avoid redundancy.
In Figure~\ref{f-biased_sampling}, we can see a comparison between uniform sampling (first row) and this curvature-based sampling (second row). Note how this sampling strategy enhanced convergence during training.

\paragraph{Image-based sampling}
For the image-based reconstruction case, the sampling for the \(\mathcal{L}_{\text{data}}\) term consists of choosing points \(\{p_i\}\) along each ray \(r(t) = p + tv\), where \(p\) is the pixel position and \(v\) is the view direction. This approach is used to discretize the volume rendering equation \cite{mildenhall2020nerf}.
Most methods follow NeRF, which splits the ray domain interval \([t_n, t_f]\) into evenly spaced times \(\{t_i\}\). Then, a time \(t'_i\) is uniformly chosen within each interval \([t_i, t_{i+1}]\).
There are numerous methods focused on improving the sampling strategy for NeRFs \cite{li2023nerfacc, barron2021mip}.

\subsection{Inference}

\begin{figure}[h!]
    \centering
    \begin{tabular}{cc}
       \includegraphics[width=0.5\columnwidth]{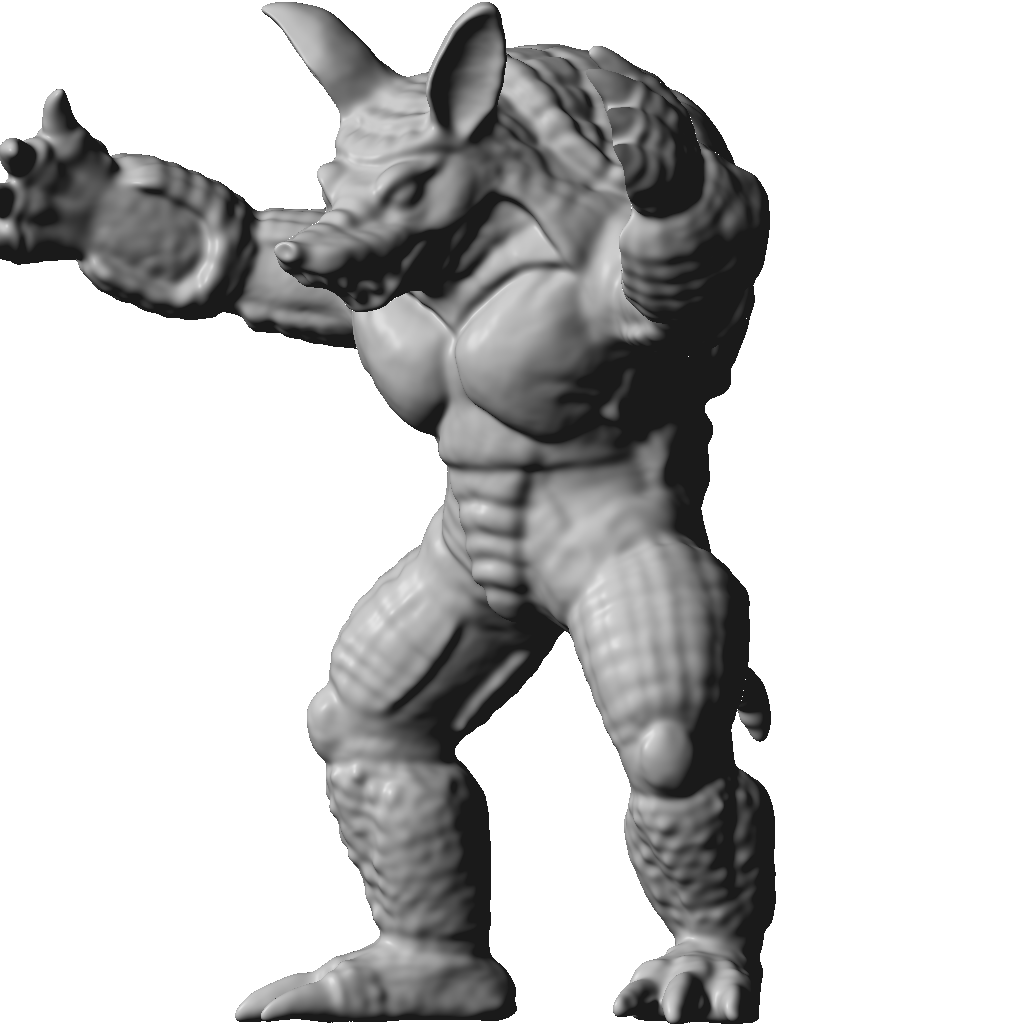}
        &
       \includegraphics[width=0.45\columnwidth]{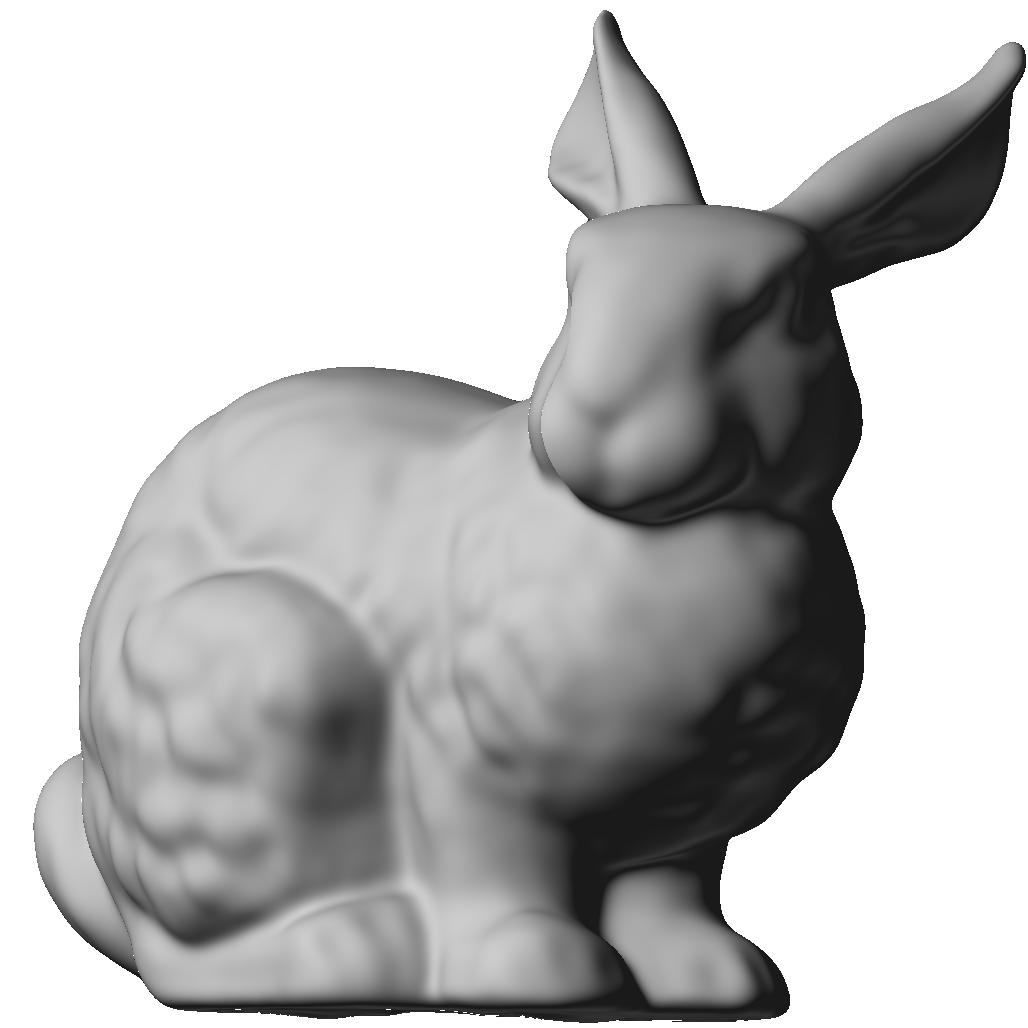}
    \end{tabular}
    \caption{Sphere tracing of neural SDFs representing the Armadillo and Bunny models. Both INRs have the same architecture and were trained on the same data during $500$ epochs. Image adapted from \cite{novello2022exploring}}
    \label{f-sdf_comparison}
\end{figure}

\paragraph{Sphere tracing}
One important advantage of neural SDFs is in rendering, as we can use the \textit{sphere tracing} algorithm~\cite{hart1996sphere}. It approximates the intersection between a ray \(r(t) = p_0 + tv\), with \(p_0\) being the starting point, and the zero-level set of a neural SDF \(f\) by iterating the system \(p_{i+1} = p_i + f(p_i)v\).
Note that this requires multiple inferences during rendering. The challenge of operating this algorithm in real-time was addressed in \cite{silva2022mip}. Figure~\ref{f-sdf_comparison} provides the sphere tracing of the zero-level set of neural SDFs representing the Armadillo and Bunny.
The algorithm can accurately ray-cast the surface, avoiding spurious components. Finally, for shading the surface, we compute the normals by simply evaluating the gradient \(\nabla f\).

\begin{figure}[h!]
    \centering
        \includegraphics[width=0.8\columnwidth]{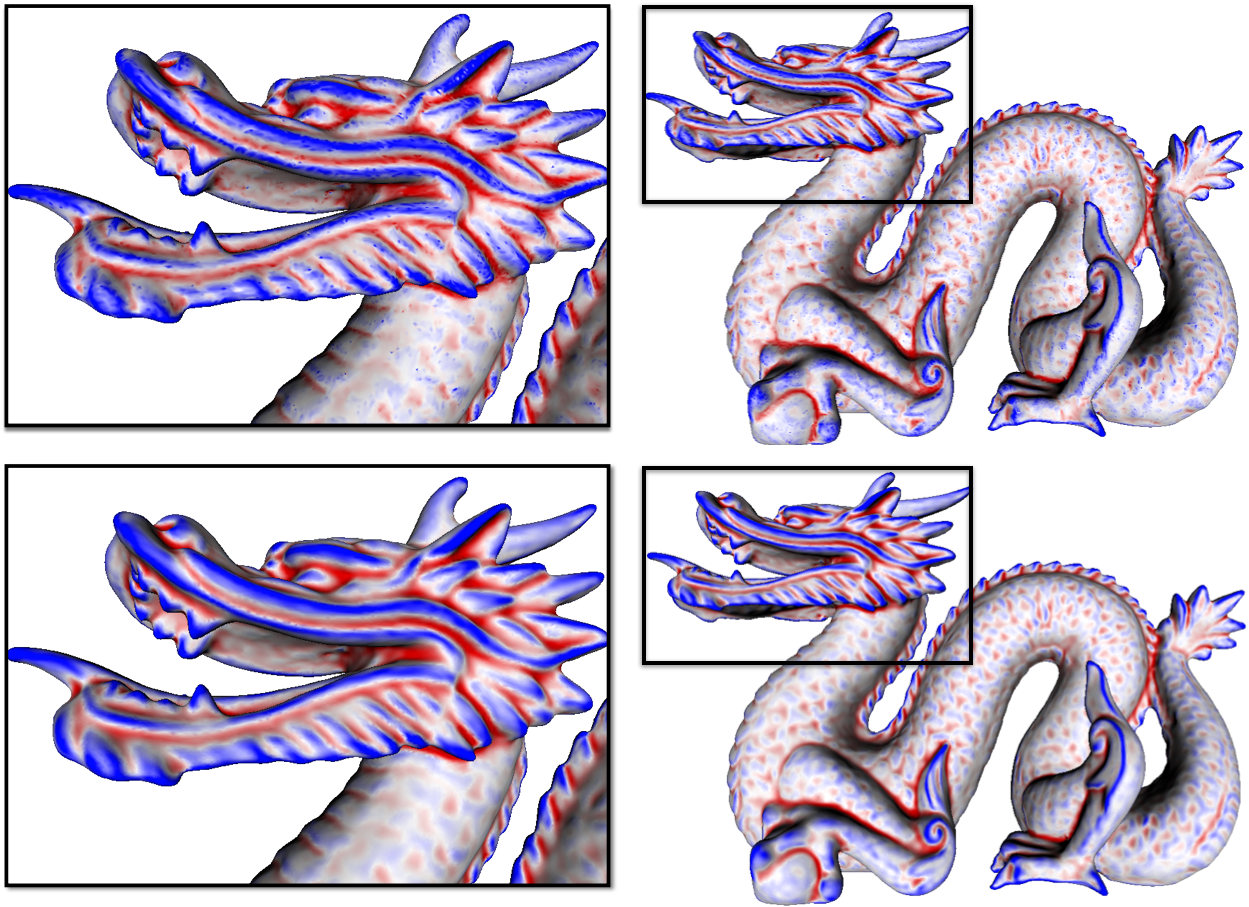}
    \caption{Visual comparison of the discrete and continuous mean curvatures of the Dragon model. The top row shows the discrete mean curvature, while the bottom row shows the mean curvatures computed from the neural SDF using \texttt{PyTorch} framework. Image from \cite{novello2022exploring}}
    \label{f-dragon}
\end{figure}

\begin{figure}[h!]
    \centering
        \includegraphics[width=0.49\columnwidth]{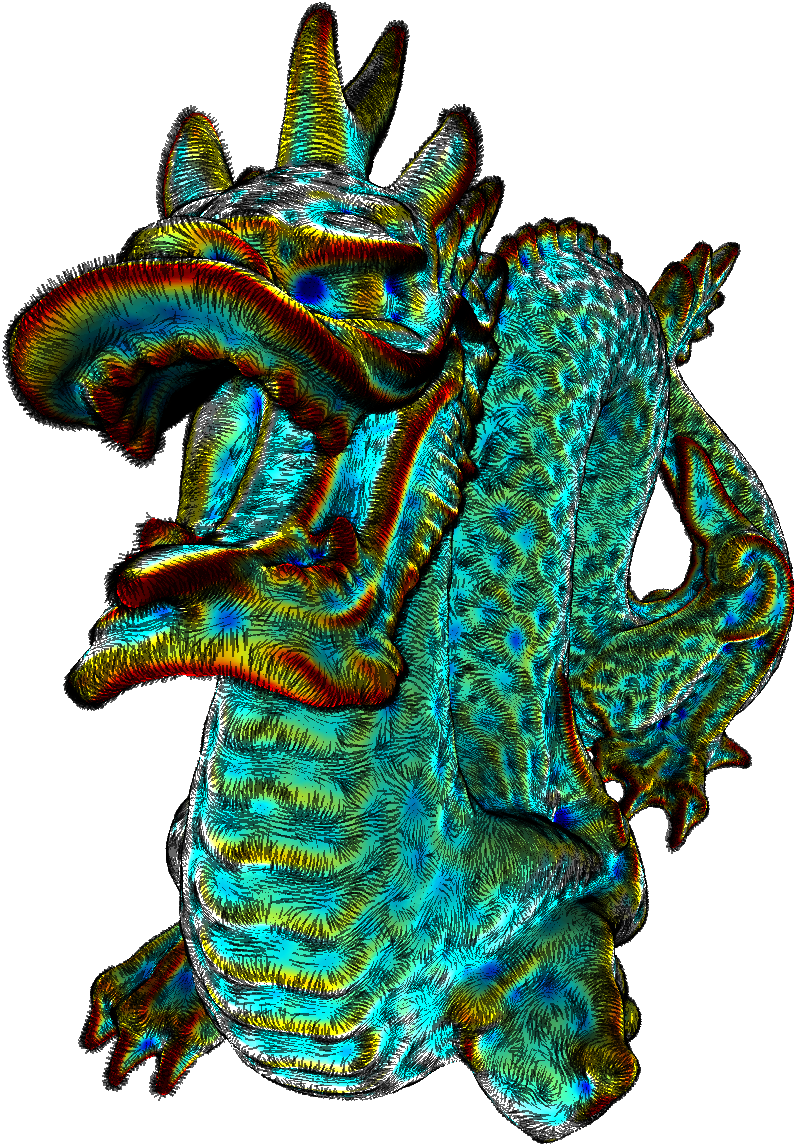}
        \includegraphics[width=0.49\columnwidth]{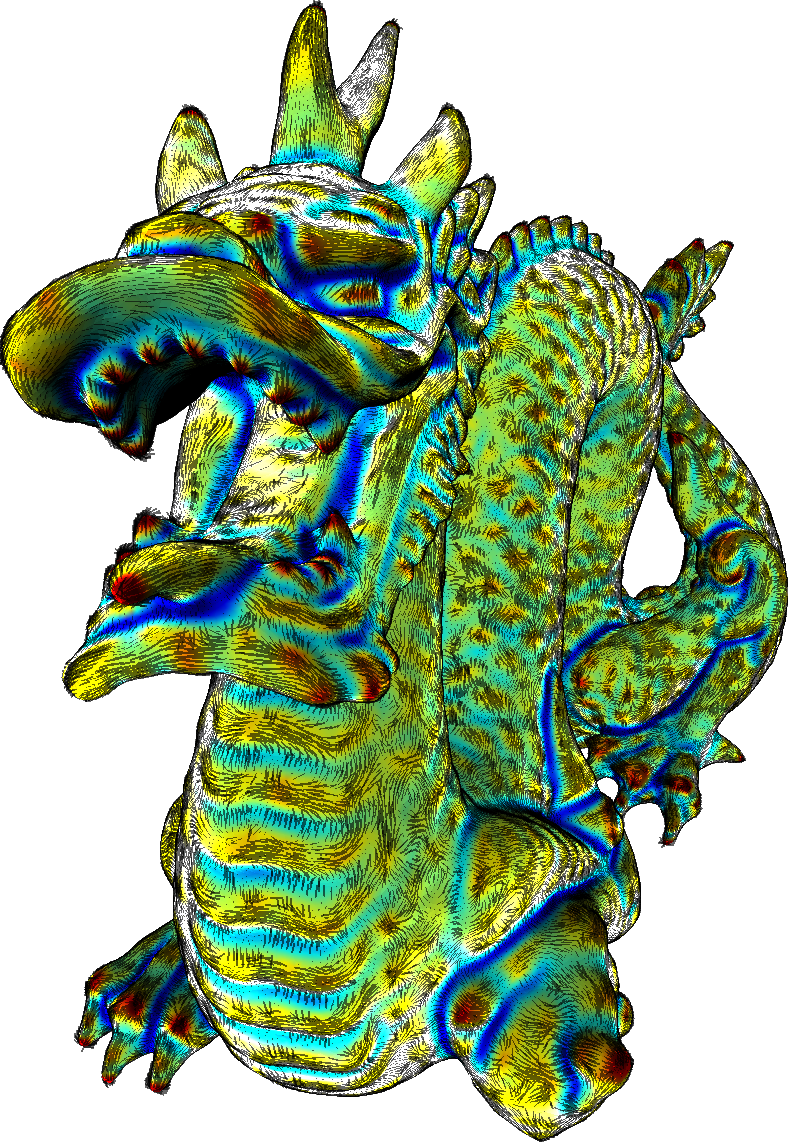}
    \caption{Principal curvatures and directions of the Dragon. Maximal curvatures are shown on the left, while minimal curvatures are on the right. Notice how their directions align nicely with the mesh's ridges and valleys.}
    \label{f-principal-armadillo}
\end{figure}

\paragraph{Curvature estimation}
Another advantage of a neural SDF \(f\) is that we can compute the curvatures of its level sets analytically, as automatic differentiation gives its second partial~derivatives. To illustrate an application, consider the input data \(\{p_i, N_i\}\) sampled from a triangle mesh representing the underlying surface \(S\). \citet{novello2022exploring} proposed transferring the curvatures of the level sets of \(f\) to the vertices of the triangle mesh.
In Figure~\ref{f-dragon}, we trained a neural SDF for the Dragon model and calculated the mean curvature using \(\Delta f\); blue indicates higher curvatures, and red indicates low curvatures. Additionally, in Figure \ref{f-principal-armadillo}, we show the principal curvatures and directions.
The study of discrete curvatures of triangle meshes is a significant topic in discrete differential geometry.

\section{Applications for INRs}
\label{s-implicits}
This section reviews a list of methods and applications of geometric INRs. We divided it in oriented point-cloud methods (Sec.~\ref{sec:app-oriented-pc}), Image-based reconstruction (Sec.~\ref{sec:nerf_based}), Multiresolution INRs (Sec.~\ref{sec:app-mr}), and Dynamic INRs (Sec.~\ref{sec:app-dynamic}).

\subsection{Surface reconstruction from oriented point clouds}
\label{sec:app-oriented-pc}

\begin{figure}[h!]
    \centering
    \includegraphics[width=1\columnwidth]{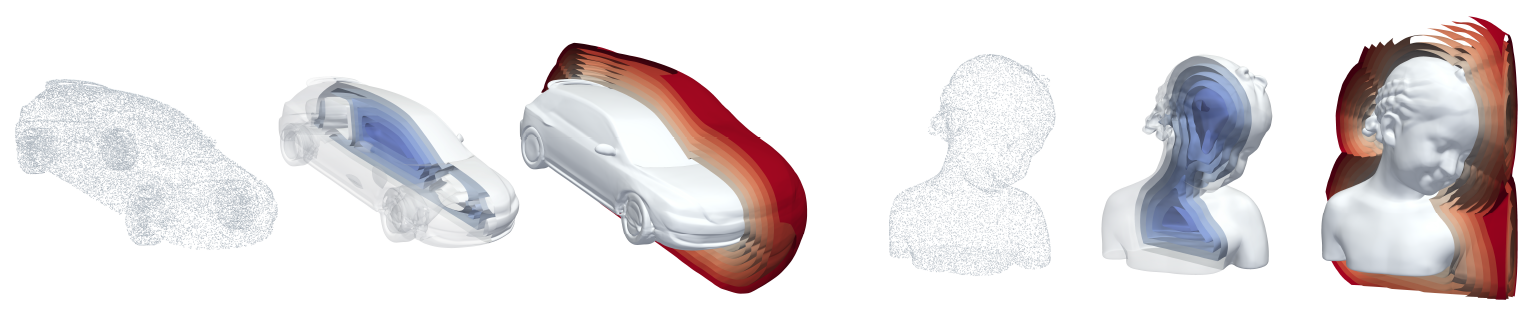}
    \caption{Level sets of MLPs trained with IGR method. Image adapted from~\cite{gropp2020implicit}.}
    \label{f-IGR}
\end{figure}

DeepSDF~\cite{park2019deepsdf} introduced SDFs in the task of representing surfaces as level sets of neural networks. However, it does not incorporate any geometric regularization in the loss function, such as enforcing the Eikonal equation~\eqref{e-eikonal}. Such regularizations were later introduced by IGR~\cite{gropp2020implicit} and SIREN~\cite{sitzmann2020implicit} which represent the underlying surface using a neural INR \(f\) and enforce the Eikonal equation during training. They consider as input an oriented point cloud \(\{ p_i, N_i \}\), thus, the data term in the loss function simply asks for \(f(p_i)=0\) and \(\nabla f(p_i) = N_i\). Figure~\ref{f-IGR} shows some level sets of MLPs trained using IGR.

\begin{figure}[ht]
    \centering
    \includegraphics[width=\columnwidth]{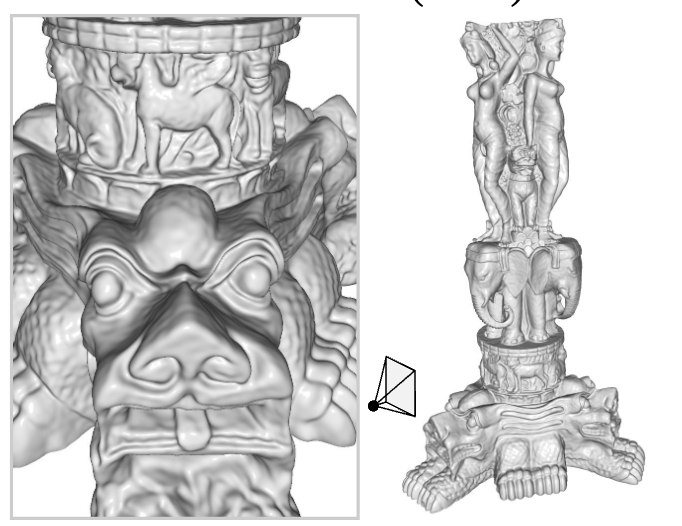}
    \caption{Siren Surface Reconstruction. Figure adapted from~\cite{sitzmann2020implicit}}.
    \label{f-siren}
\end{figure}

The major difference between IGR and SIREN lies in the architecture of the INR. SIREN proposes parameterizing the INR using a sinusoidal MLP (\(\varphi = \sin\) in Equation~\ref{e-network-architecture}) with an initialization scheme that ensures stability and good convergence during training. Due to their smoothness and large representation capacity, SIRENs have emerged as one of the most popular architectures, influencing many other works in surface representation~\cite{novello2022exploring, sundt2023marf, jiao2023naisr, fainstein2024dudf}. Figure~\ref{f-siren} shows a surface reconstruction using SIREN to fit an SDF from an oriented point cloud.

\citet{novello2022exploring} introduced a geometric INR which explores the curvature of the data during the sampling stage to speed up training by selecting regions with more geometric details. Section~\ref{s-sampling} provide additional details. 
This method also proposes a loss function that incorporates the curvature of the INR level sets. For this, it uses the closed-form derivatives of the network to estimate differential measures, such as normals and curvatures. This estimation is feasible because the oriented point cloud lies in the neighborhood of the INR zero-level set.
Later, \citet{jiao2023naisr} applied this approach to 3D shape reconstruction and analysis of real medical data. They leveraged geometric INRs to create a shape atlas that captures the effects of age, sex, and weight on 3D shapes, enabling shape reconstruction and evolution.

Representing surfaces using SDFs has a common limitation: they can only represent closed surfaces. This is because SDFs inherently assume a clear distinction between the ``inside" and ``outside" of the underlying surface, making them less suitable for open surfaces. \citet{fainstein2024dudf} addressed this limitation by extending the approach in \cite{novello2022exploring} to open surfaces using \textit{unsigned} distance functions (UDFs).

\subsection{Neural Implicit Surface Reconstruction from Images}
\label{sec:nerf_based}

IDR~\cite{yariv2020idr} is one of the first geometric INR approaches to learn a neural SDF from images. IDR simultaneously learns the neural SDF, camera parameters, and a neural renderer that approximates the light reflected from surfaces toward the camera. The geometry is represented as the zero-level set of an MLP, while a neural renderer, based on the rendering equation, implicitly models lighting and materials.
Volume SDF~\cite{yariv2021volume} adopts a different approach by using \textit{differentiable volume rendering} to optimize the INR parameters. It maps the SDF to a volume density function, assigning a value of 1 to points on the zero-level set and 0 to points far from this set. The neural SDF is optimized through \textit{volume rendering} from the images and is regularized using the Eikonal constraint.

\begin{figure}[ht]
    \centering
    \includegraphics[width=\columnwidth]{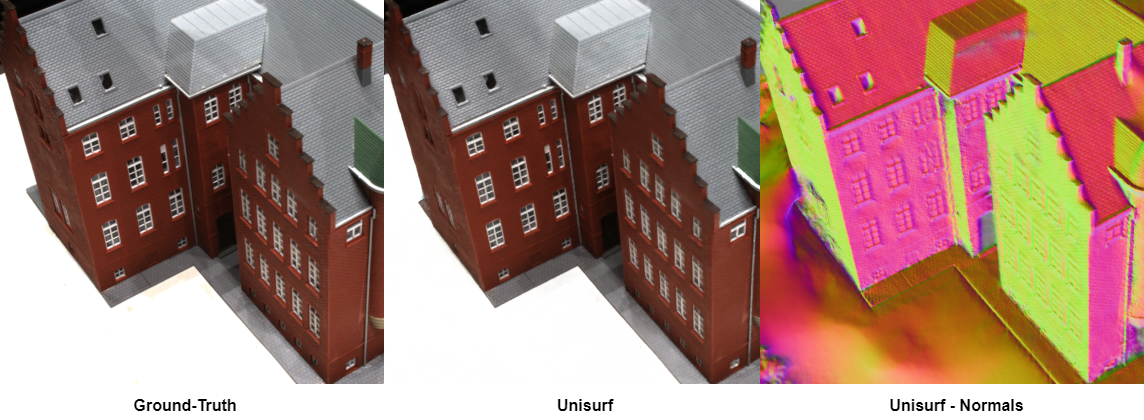}
    \caption{Reconstruction of scan 24 of DTU dataset evaluated with UNISURF.}
    \label{f-unisurf}
\end{figure}

Exploring volume rendering in 3D reconstruction has been motivated by the groundbreaking results of \textit{Neural radiance fields} (NeRF)~\cite{mildenhall2020nerf, yu2021pixelnerf, zhang2020nerf++, barron2021mip}. NeRF employs differentiable volume rendering to optimize the parameters of a \texttt{ReLU} MLP combined with a \textit{Fourier feature mapping}~\cite{tancik2020fourier}. These networks are trained using volume rendering with supervision from a set of posed images. In this approach, surface geometry is represented as a density function, where the value is \(1\) for points on the surface and \(0\) for points far from it. However, surfaces extracted from such neural densities (using marching cubes~\cite{lorensen1998marching}) tend to be noisy (see Figure~\ref{f-unisurf}).

To address the noise issues in NeRF-based surface extraction, \citet{Oechsle2021ICCV} proposed UNISURF. This method improves the reconstruction quality by replacing NeRF's density function with an \textit{occupancy network}~\cite{mescheder2019occupancy}, offering a more robust representation of surface geometry.
%
Figure~\ref{f-unisurf} shows their method evaluated on the DTU dataset \citet{jensen2014large}.

However, UNISURF does not leverage Eikonal regularization because its underlying model is not an SDF. To our knowledge, NeuS~\cite{wang2021neus} is the first geometric INR method to adapt the NeRF pipeline for training neural SDFs from posed images. It employs a density distribution function (see Sec~\ref{sec:loss}) to map SDF values to density values.

\begin{figure}[ht]
    \centering
        \includegraphics[width=\columnwidth]{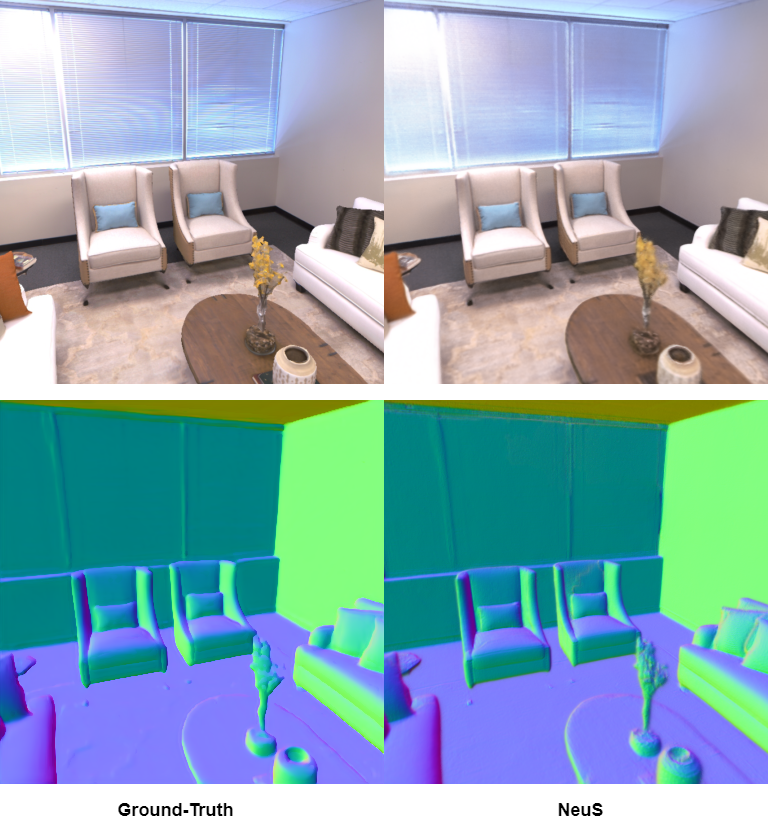}
    \caption{ Results from NeuS on the Room scene from the dataset available in the paper Neural RGB-D Surface Reconstruction\cite{azinovic2022neural}. The figure shows the scene reconstruction as well as the normal map in comparison to the input data. In these experiments, we ran the NeuS method with its default settings for 20,000 steps.
    }
    \label{f-neus}
\end{figure}

NeuS motivated several techniques, addressing aspects such as incorporating depth maps~\cite{yu2022monosdf}, handling sparse views~\cite{long2022sparseneus}, baking the neural SDF~\cite{yariv2023bakedsdf}, enhancing gradient consistency~\cite{ma2023towards}, patch warping (NeuralWarp)~\cite{darmon2022improving}, and applying curvature regularization (Neuralangelo)~\cite{li2023neuralangelo}. Figure~\ref{f-neus} shows results from NeuS on the room scene from the dataset presented by ~\citet{azinovic2022neural}.

Neural RGB-D Surface Reconstruction~\cite{azinovic2022neural} presents a method for 3D scene reconstruction that effectively combines RGB and depth data using a hybrid scene representation based on a truncated signed distance function (TSDF) and a volumetric radiance field. The authors address limitations of traditional methods that struggle with noisy depth measurements and incomplete geometry by leveraging color information to fill in gaps where depth data is lacking. Their approach not only improves the quality of geometry reconstructions but also optimizes camera poses to reduce misalignment artifacts, demonstrating superior performance compared to methods like NeRF, NeuS and UNISURF with depth constraints, particularly in complex indoor environments.

\begin{figure}[!ht]
    \centering
    \includegraphics[width=\columnwidth]{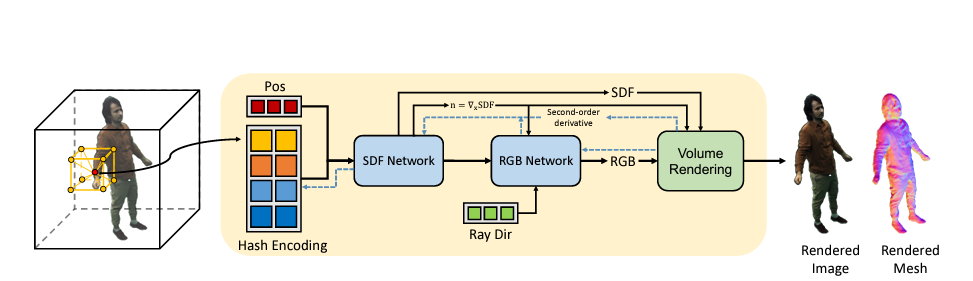}
    \caption{ Given a point $p$, NeuS2 combine its feature from a hashgrid with its coordinates as the input for the neural SDF, which also outputs geometric features. These are subsequently combined with the viewing direction and fed into a RGB network to produce the color value. Figure adapted from \cite{wang2023neus2}.}
    \label{f-neus2}
\end{figure}

Next, NeuS2~\cite{wang2023neus2} and InstantNSR~\cite{zhao2022human} used the the hashgrid-based network architecture of instant-NGP~\cite{muller2022instant} to make training and rendering faster. In the same vein, Neuralangelo~\cite{li2023neuralangelo} also uses hashgrids mixed with finite gradients methods to make training have a much better performance. Figure \ref{f-neus2} presents an overview of the NeuS2 architecture.

\begin{figure}[ht]
    \centering
    \includegraphics[width=\columnwidth]{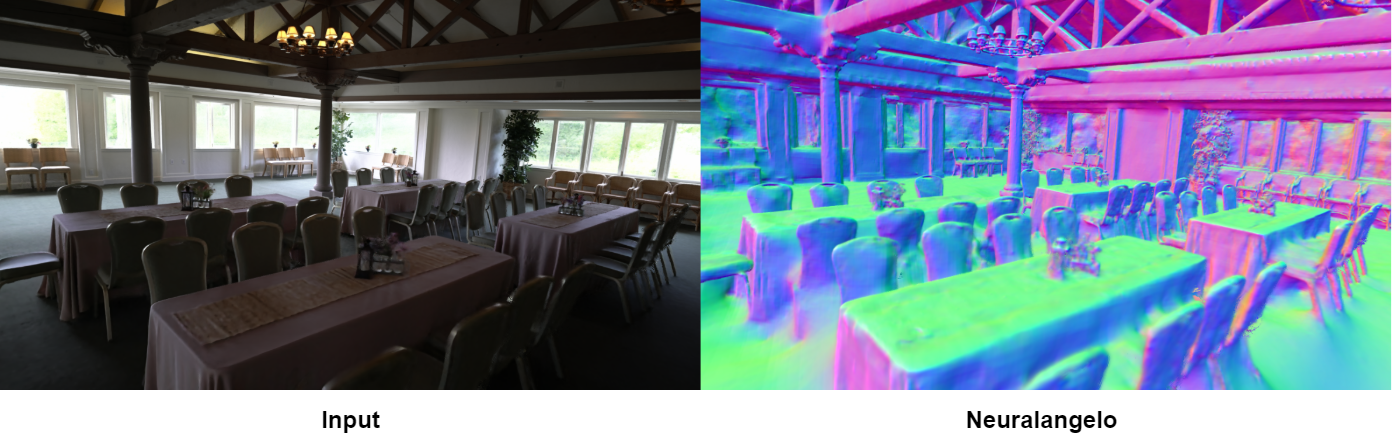}
    \caption{Qualitative comparison on Tanks and Temples dataset\cite{knapitsch2017tanks}.
    }
    \label{f-neuralangelo}
\end{figure}

Table~\ref{tab:sdf_nerf}, shows a comparison between some techniques on the scene 122 of DTU dataset, both in the view synthesis task and the geometry reconstruction.
Neuralangelo~\cite{li2023neuralangelo} is, currently, the state-of-the-art in both tasks. 
It is a example of geometric INR since it parameterizes the underlying surface as a neural SDF and includes the mean curvature of the level sets as a regularizer. The tests were conducted on a computer with an AMD Ryzen 7 5700 CPU, 32 GB of RAM, and an NVIDIA RTX 3080 graphics card.
Figure~\ref{f-neuralangelo} shows a qualitative evaluation with Neuralangelo using the Tanks and Temples dataset\cite{knapitsch2017tanks}.
\begin{table}[!ht]
\centering
\label{tab:tab1}
\begin{tabular}{lrr}
\toprule
\textbf{Method}     & \textbf{Mean PSNR ($\uparrow$)} & \textbf{Chamfer dist.} ($\downarrow$)    \\ \midrule
NeuS   & 30.1             &   0.56  \\


VolSDF        &  30.38               &   0.58      \\

UNISURF        &   27.32              & 0.66    \\

Neuralangelo         & \textbf{34.91}               & \textbf{0.45}     \\ \bottomrule
\end{tabular}
\caption{
Performance of Neural SDF methods on the scene 122 of DTU dataset. The PSNR measures the quality of the synthesized view while the Chamfer distance measures the quality of the geometry of the SDF by computing the Chamfer distance for the synthesized mesh.}
\label{tab:sdf_nerf}
\end{table}

Other than NeuS2, other techniques tried to improve the speed of these representations.
For example, MARF~\cite{sundt2023marf} introduce the Medial Atom Ray Fields, a neural object representation enabling differentiable surface rendering with a single network evaluation per camera ray.
MARFs address challenges like multi-view consistency and surface discontinuities by using a medial shape representation, offering cost-effective geometrically grounded surface normals and analytical curvature computation. They map camera rays to multiple medial intersection candidates and demonstrate applicability in sub-surface scattering, part segmentation, and representing articulated shapes. With the ability to learn shape priors, MARFs hold promise for tasks like shape retrieval and completion.

\begin{figure}[ht]
    \centering
    \includegraphics[width=\columnwidth]{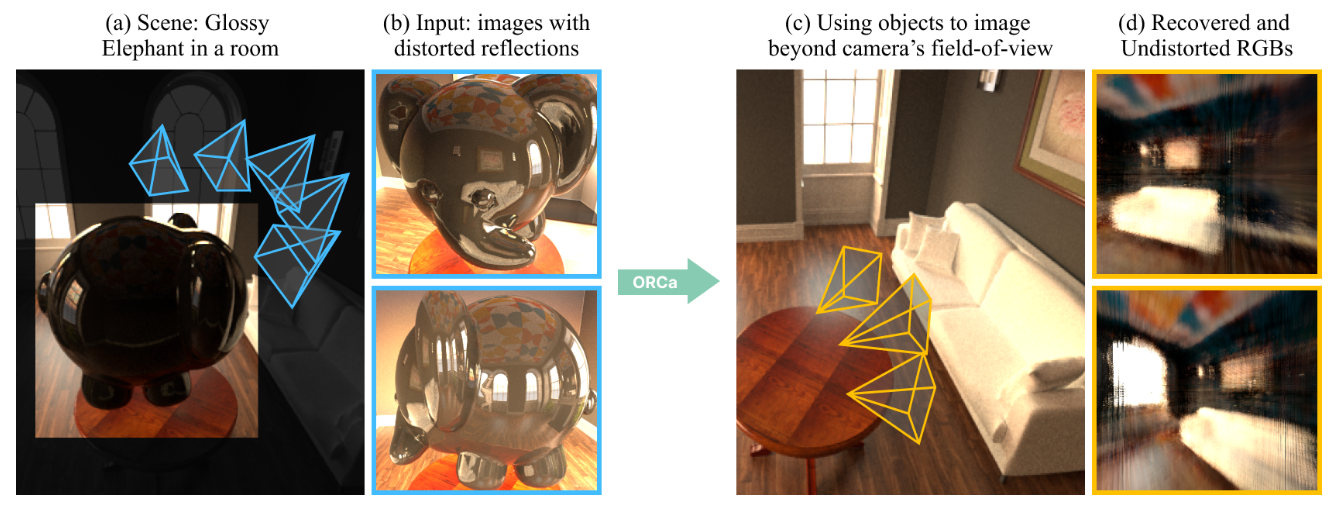}
    \includegraphics[width=\columnwidth]{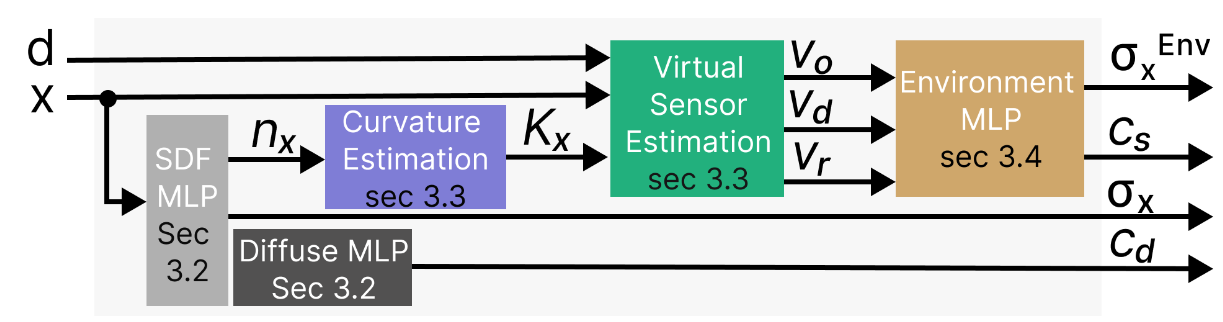}
        \caption{ The diagram presents results from the ORCa model and its inputs. It converts objects with unknown geometry into radiance-field cameras by modeling multi-view reflections as projections of a 5D radiance field. Additionally, it transforms the object surface into a virtual sensor to capture this radiance field, enabling depth and radiance estimation of the surrounding environment. The model then queries this radiance field to perform novel view synthesis beyond the field-of-view. ORCa consists of three steps: modeling the object’s geometry as a neural implicit surface, converting the object’s surface into a virtual sensor, and projecting the environment’s radiance field along these virtual cones. The learned radiance field of the environment allows visualization of occluded areas in novel viewpoints. Figure adapted from~\cite{tiwary2023orca}.
        }
    \label{f-orca_diag}
\end{figure}

Recent research on recovering scene properties from images often employ neural SDFs and differential geometry, with deep neural networks demonstrating inverse rendering of indoor scenes from a single image. However, these methods typically yield coarse lighting representations and lack fine environmental details. Approaches like Lighthouse \cite{srinivasan2020lighthouse} and NeRFFactor \cite{zhang2021nerfactor} train on natural illumination maps to address environment estimation challenges. PANDORA \cite{dave2022pandora} and RefNeRF utilize multi-view reflections but assume a distant environment modeled with a flat 2D map, while PhySG \cite{zhang2021physg} models surfaces as SDFs for multi-view image rendering. ORCa \cite{tiwary2023orca}, building on \citet{novello2022exploring}'s concepts, uses reflections on glossy objects to capture hidden environmental information by converting these objects into radiance-field cameras. This transforms object surfaces into virtual sensors, capturing reflections as 2D projections of the 5D environment radiance field. This technique enables depth and radiance estimation, novel-view synthesis beyond the field of view, and imaging around occluders, using differential geometry to estimate curvature for neural implicit surfaces. Figure \ref{f-orca_diag} shows the results of this method and a diagram of its architecture.

To test these different types of SDF reconstruction techniques, and inspired by nerfstudio~\cite{nerfstudio}, \citet{Yu2022SDFStudio} created SDFStudio, a platform that allows the testing and prototyping of different SDF extraction techniques from images.
Table \ref{tab:tab2} show a comparison between the methods, following the input, geometry regularization methods employed.

\begin{table*}[!ht]
\centering
\label{tab:tab2}
\begin{tabular}{lll}
\toprule
\textbf{Method}     & \textbf{Input}                        & \textbf{Geometric regularization}            \\ \midrule
PANDORA~\cite{dave2022pandora}      & Multi-view and polarized RGB & Eikonal Constraint          \\
PhySG~\cite{zhang2021physg}        & Multi-View RGB               & Eikonal Constraint         \\
ORCa~\cite{tiwary2023orca}         & Multi-View RGB               & Eikonal Constraint, Mean Curvature Constraint       \\
NeuS~\cite{wang2021neus}         & Multi-View RGB               & Eikonal Constraint,                         \\
NeuS2~\cite{wang2023neus2}        & Multi-View RGB               & Eikonal Constraint                      \\
Neuralangelo~\cite{li2023neuralangelo} & Multi-View RGB               & Eikonal Constraint, Mean Curvature          \\
Neural Warp~\cite{darmon2022improving} & Multi-View RGB               & Eikonal Constraint         \\
Geo-NeuS~\cite{fu2022geo} & Multi-View RGB              & Eikonal Constraint         \\
VolSDF~\cite{yariv2021volume} & Multi-View RGB              & Eikonal Constraint          \\
MonoSDF~\cite{yu2022monosdf} & Multi-View RGB, Depth/Normals               & Eikonal Constraint, Monocular Supervision          \\ \bottomrule
\end{tabular}

\caption{
    Breakdown of SDF Reconstruction from multi-view images. We classify them based on the type of input and geometric regularization approaches.}
\end{table*}

\subsection{Multiresolution INRs}
\label{sec:app-mr}

\begin{figure}[h!]
    \centering
        \includegraphics[width=0.8\columnwidth]{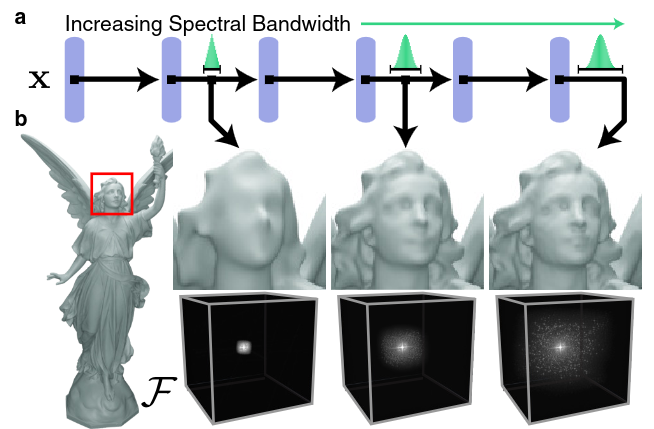}
        \caption{The BACON architecture generates intermediate results with a specific spectral bandwidth chosen during initialization. When trained with high-resolution data, the network learns to decompose outputs across multiple resolutions, useful for tasks like fitting 3D shapes using signed distance functions. The characteristics of the network are defined by its Fourier spectrum, ensuring constrained behavior even in unsupervised scenarios. Figure adapted from \cite{lindell2022bacon}.
        }
    \label{f-bacon}
\end{figure}

Multiresolution is a well-studied concept in classical geometry processing~\cite{mederos2003moving}.
Given this, and inspired by traditional multi-resolution theory for images~\cite{williams1983pyramidal, mallat1989theory}, \citet{lindell2022bacon} introduce BACON, an architecture aimed at addressing the confinement to single-scale signal representation for coordinate-based networks.
BACON proposes the creation of an analytical Fourier spectrum, enabling controlled behavior at unsupervised points and allowing for design driven by the signal of spectral characteristics. This method also supports multiscale signal representation without needing supervision at every scale. It has been applied to the neural representation of images, radiance fields, and 3D scenes using SDFs, demonstrating its effectiveness in representing signals across various scales. Figure \ref{f-bacon} shows an overview of the method.

In addressing the limitations of BACON, it is noted that its cut-off of the Fourier spectrum introduces artifacts, particularly the ringing effect observed in images and noise on surfaces. This limitation stems from the inherent constraints of band-limiting, which, while facilitating certain advantages in signal representation, can also result in undesirable visual artifacts that affect the quality of the output in applications involving high-frequency detail.
Other techniques such as BANF~\cite{shabanov2024banf}, build on BACON, but does so with filtering during optimization, which allows it to have a band-limited frequency decomposition.
This allows our application of SDF, to have an SDF in multiple levels of detail.

Other techniques allow numerous level-of-detail for rendering different images, for example \citet{takikawa2021neural}, allow for rendering in multiple levels-of-detail and MINER~\cite{saragadam2022miner} allows the training on multiple levels of scale.

\subsection{Dynamic INRs}
\label{sec:app-dynamic}



INRs serve as efficient and versatile geometry representations, encoding both functional and differential data of the underlying object compactly. However, they lack intuitive control over shape editing and animations. Here we explore methods to manipulate implicit surfaces by deformations, whether rigid or not, shape interpolation, and animations. We also list interesting examples of works that do not strictly fit the definition of INR, but exploit differential properties of smooth networks in their objective function.

\begin{figure}[ht]
    \centering
    \includegraphics[width=\columnwidth]{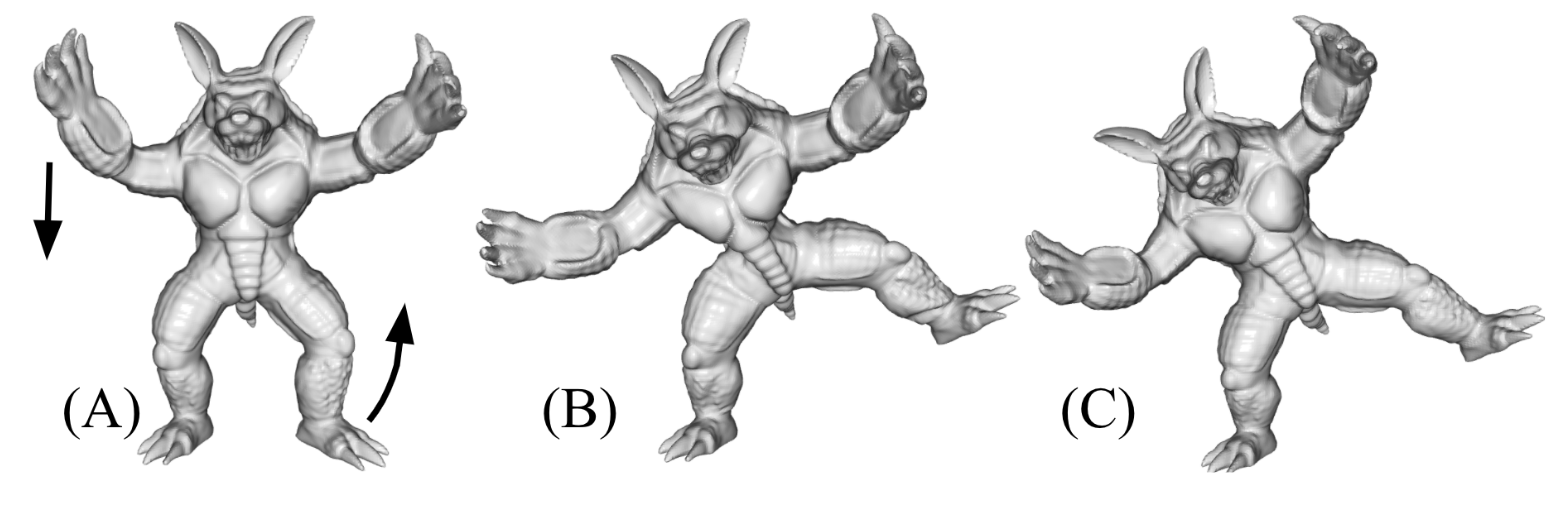}
    \caption{Deformation results from \citet{yang2021geometry}: (A) Input shape, (B) Baseline and (C) their method. Their method applies multiple transformation on the Armadillo. Figure adapted from \cite{yang2021geometry}.}
    \label{f-nfgp}
\end{figure}

\citet{niemeyer2019occupancy} presents the first approach to leverage deep learning for 3D surface animation. \emph{Occupancy flow} extends Occupancy Networks~\cite{mescheder2019occupancy} by learning a vector field in addition to the occupancy for each point in $R^3$ continuously along time, thus animating objects in the scene. Their work achieves good results without discretizations and shape templates, both usual techniques incorporated in most contemporaneous methods.
More recently~\citet{yang2021geometry} proposed to leverage INRs for geometry processing tasks, specifically smoothing and sharpening, in addition to more complicated rigid deformations, such as twists and bends. The authors do not propose their method for animation specifically, but geometrical deformations, thus we classify them under the umbrella of dynamic INRs. The authors propose approximating a local surface of a level set by utilizing the derivatives of the underlying field. By solely relying on the field derivatives, it is possible to use intrinsic geometric properties of the level set, such as curvatures. This enables the construction of loss functions that capture surface priors like elasticity or rigidity. This is made possible by exploiting the inherent infinite differentiability of specific neural fields which facilitates the optimization of loss functions involving higher-order derivatives through gradient descent methods. Consequently, unlike mesh-based geometry processing algorithms that rely on surface discretizations to approximate these objectives, this strategy can directly optimize the derivatives of the field. Their method can apply transformation like rotation or translation on complex 3D objects as we can see in Figure \ref{f-nfgp}.

Similarly to \citet{yang2021geometry}, \citet{mehta2022level} proposes the use of sinusoidal MLPs to apply smooth deformations, smoothing, and sharpening to surfaces parameterized as INRs.
However, as in \citet{yang2021geometry} they still need supervision during the intermediate time-steps to learn a deformation of the base surface. This supervision is done by converting the implicit object to an explicit representation via marching cubes. While their work achieves good results, it does not achieve smoothness along time, since each time-step irreversibly modifies the INR, making it impossible to walk through the deformation after training.

\begin{figure}[ht]
    \centering
    \includegraphics[width=\columnwidth]{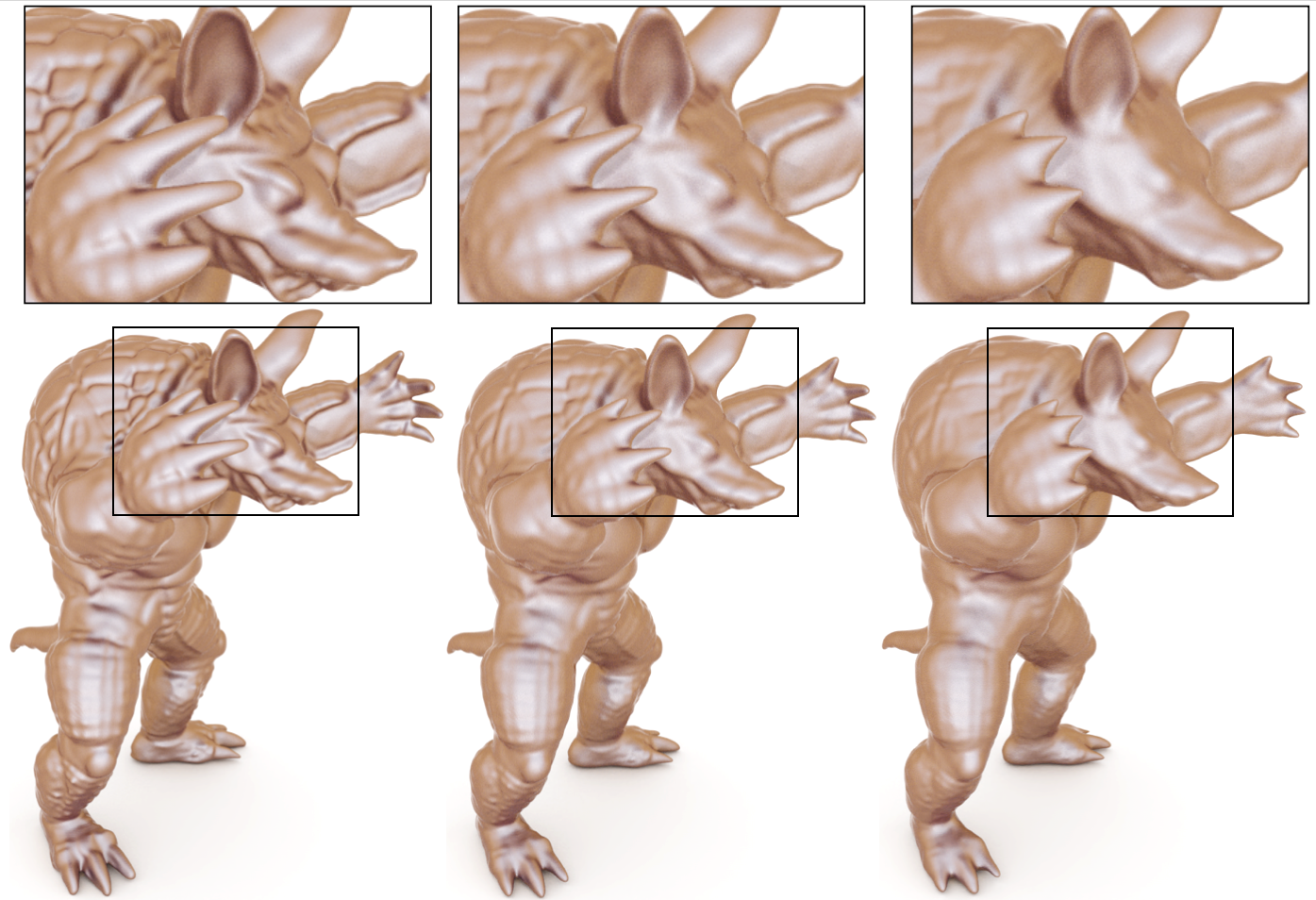}
    \includegraphics[width=\columnwidth]{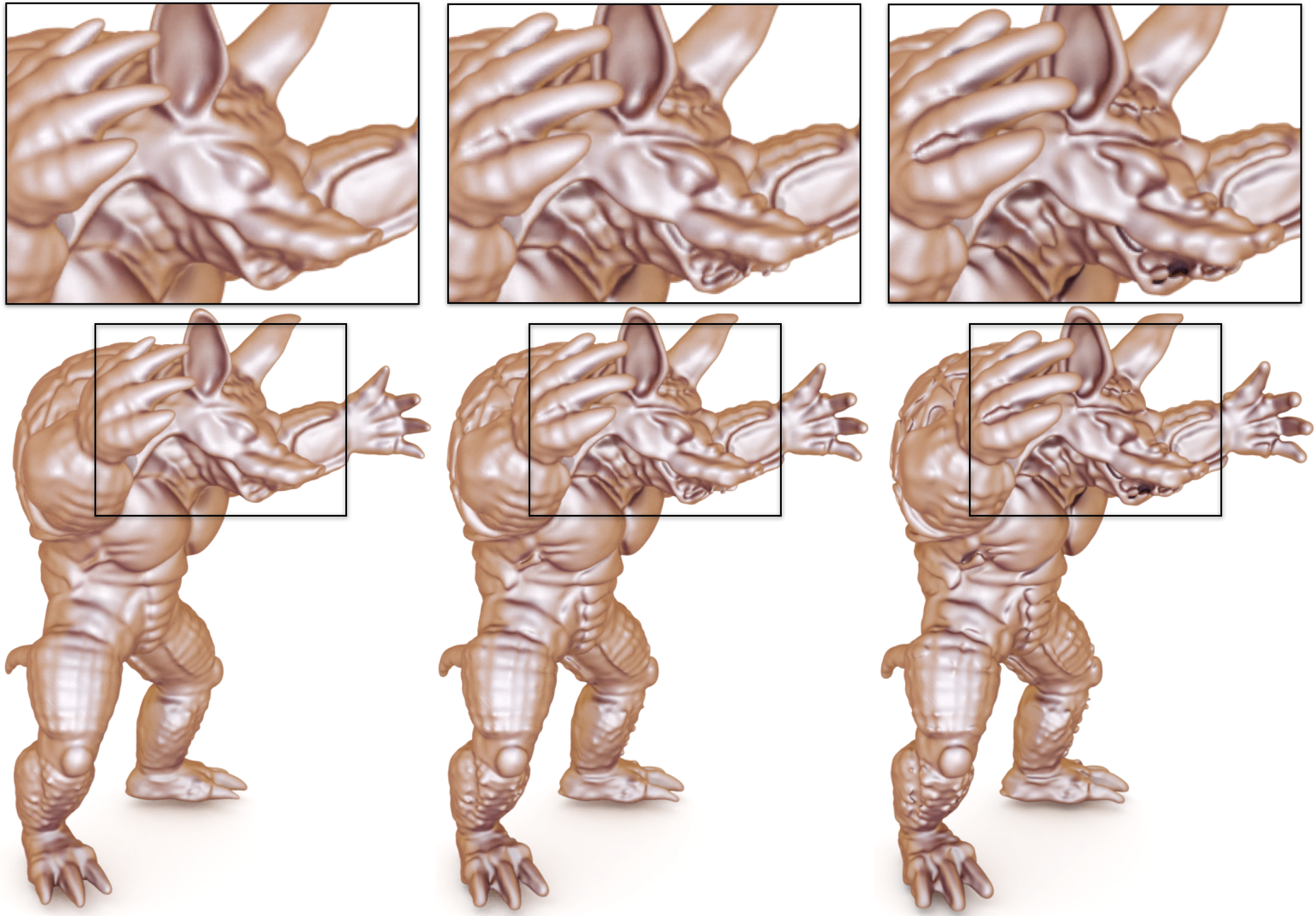}
    \caption{Smoothing and Sharpening the Armadillo with the method from \citet{novello2023neural}.}
    \label{f-i4d}
\end{figure}

\citet{novello2023neural} proposes to incorporate the differential equation directly into the loss function, thus removing the need for intermediate-time discretizations. As in~\cite{niemeyer2019occupancy}, \citet{novello2023neural} expands the domain to include the time parameter ($f:\R^3\times \R\rightarrow \R$ instead of $f:\R^3\rightarrow \R$), resulting in a smooth representation in both space and time. The authors proposed to leverage the Mean Curvature Flow for smoothing and sharpening of the base INR, similarly to \cite{yang2021geometry} and \cite{mehta2022level} as we can see in Figure \ref{f-i4d}.

\begin{figure}[ht]
    \centering
    \includegraphics[width=\columnwidth]{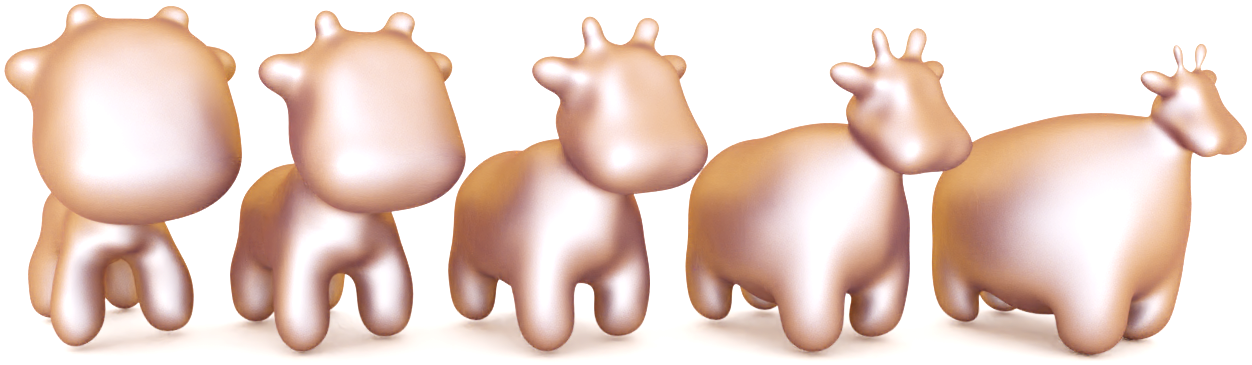}
    \includegraphics[width=0.9\columnwidth]{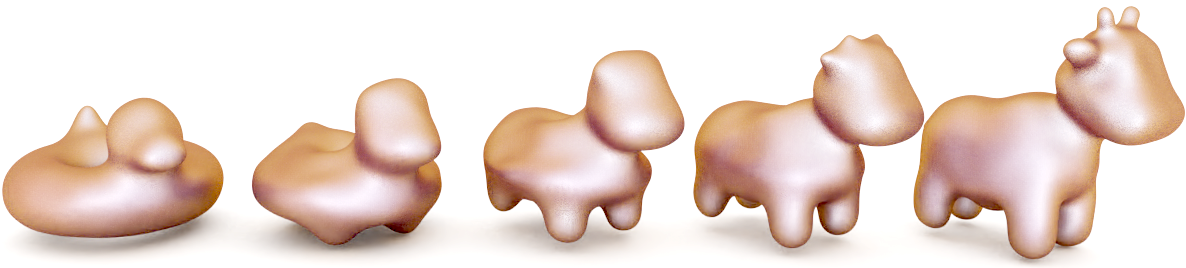}
    \caption{Deformation and interpolation of INRs. Top row shows the evolution of the zero-level sets of an INR according to a vector field with a source and a sink. The SDF of the Spot is the initial condition at $t=0$ (middle). The sink/source is inside the head/body of the Spot. Bottom row shows an example of shape interpolation between Bob, positioned at $t=0$ and Spot, at $t=1$. Images adapted from~\cite{novello2023neural}.}
    \label{f-spot_vector_field}
\end{figure}

Additionally, they propose to exploit the Level Set equation for shape interpolation, as in \cite{mehta2022level}, achieving fully continuous deformations parameterized by a single neural network. See Fig.~\ref{f-spot_vector_field} for interpolation and deformation examples.

An interesting application of dynamic INRs is on the face-morphing task, where the goal is to create a transition between two or more existing faces, or different expressions of the same face. In this context, \citet{zheng2022imface} proposes to leverage INRs by creating expression and identity deformations between previously captured faces. They expand on DeepSDF~\cite{park2019deepsdf}, and create two separate latent spaces: one for expressions , and one for identities. A third network encodes a template shape, which is the average face of the dataset. The authors further expanded their method by incorporating a refined displacement grid to capture finer facial details~\cite{zheng2023imfacepp}. Both works incorporate the Eikonal term as a regularizer for the loss function in addition to a normal alignment term.
Using a separate network for the flow, as done in another domain in \citet{schardong2024neural}, holds promise for this field.

\citet{wang2023neus2} expands NeuS~\cite{wang2021neus} to dynamic scene reconstruction from videos, by improving the training performance of NeuS with HashGrids introduced by InstantNGP~\cite{muller2022instant}. They exploit a key property that the main object remains reasonably static on successive frames and propose an incremental training scheme from a base INR built from the first video frame. Additionally, they regularize their loss term using an Eikonal term, similarly to \citet{novello2022exploring}.

\section{Conclusion}
\label{s-conclusion}
The paper reviewed INR approaches for solving the problem of implicit surface reconstruction by applying geometric regularization to the INR's level sets. To present these approaches, we introduced a \textit{geometric INR} framework, outlining a commonly used pipeline in this field. We defined the key components of this pipeline, including \textit{input data}, which can consist of either oriented point clouds or posed images, and a \textit{geometric loss function}, which leverages the differential properties of the function for regularization and sampling during training.

Additionally, we explored several applications that leverage the differentiability of neural networks and the discrete geometry of oriented point clouds to fit the INR's zero-level set to the data. Our review covers prominent geometric INRs, such as SIREN and IGR, along with state-of-the-art methods from image-based reconstruction literature, like NeuS and Neuralangelo. These methods demonstrate that incorporating differential geometry concepts into the INR loss function offers significant potential for reconstructing surfaces from posed images.

\bibliographystyle{plainnat}
\bibliography{refs}

@article{sitzmann2020implicit,
  title={Implicit neural representations with periodic activation functions},
  author={Sitzmann, Vincent and Martel, Julien and Bergman, Alexander and Lindell, David and Wetzstein, Gordon},
  journal={Advances in Neural Information Processing Systems},
  volume={33},
  year={2020}
}

@misc{sang2023enhancing,
      title={Enhancing Surface Neural Implicits with Curvature-Guided Sampling and Uncertainty-Augmented Representations},
      author={Lu Sang and Abhishek Saroha and Maolin Gao and Daniel Cremers},
      year={2023},
      eprint={2306.02099},
      archivePrefix={arXiv},
      primaryClass={cs.CV},
      url={https://arxiv.org/abs/2306.02099},
}

@InProceedings{niemeyer2019occupancy,
  author={Niemeyer, Michael and Mescheder, Lars and Oechsle, Michael and Geiger, Andreas},
  booktitle={2019 IEEE/CVF International Conference on Computer Vision (ICCV)},
  title={Occupancy Flow: 4D Reconstruction by Learning Particle Dynamics},
  year={2019},
  volume={},
  number={},
  pages={5378-5388},
  doi={10.1109/ICCV.2019.00548}
}

@article{tancik2020fourier,
  title={Fourier features let networks learn high frequency functions in low dimensional domains},
  author={Tancik, Matthew and Srinivasan, Pratul and Mildenhall, Ben and Fridovich-Keil, Sara and Raghavan, Nithin and Singhal, Utkarsh and Ramamoorthi, Ravi and Barron, Jonathan and Ng, Ren},
  journal={Advances in neural information processing systems},
  volume={33},
  pages={7537--7547},
  year={2020}
}

@book{szeliski2022computer,
  title={Computer vision: algorithms and applications},
  author={Szeliski, Richard},
  year={2022},
  publisher={Springer Nature}
}

@inproceedings{gropp2020implicit,
author = {Gropp, Amos and Yariv, Lior and Haim, Niv and Atzmon, Matan and Lipman, Yaron},
title = {Implicit geometric regularization for learning shapes},
year = {2020},
publisher = {JMLR.org},
booktitle = {Proceedings of the 37th International Conference on Machine Learning},
articleno = {355},
numpages = {11},
series = {ICML'20}
}

@inproceedings{mehta2022level,
  title={A level set theory for neural implicit evolution under explicit flows},
  author={Mehta, Ishit and Chandraker, Manmohan and Ramamoorthi, Ravi},
  booktitle={European Conference on Computer Vision},
  pages={711--729},
  year={2022},
  organization={Springer}
}

@article{novello2022exploring,
  title={Exploring differential geometry in neural implicits},
  author={Novello, Tiago and Schardong, Guilherme and Schirmer, Luiz and da Silva, Vinicius and Lopes, Helio and Velho, Luiz},
  journal={Computers \& Graphics},
  volume={108},
  pages={49--60},
  year={2022},
  publisher={Elsevier}
}

@inproceedings{schirmer2021neural,
  title={Neural Networks for Implicit Representations of 3D Scenes},
  author={Schirmer, Luiz and Schardong, Guilherme and da Silva, Vin{\'\i}cius and Lopes, H{\'e}lio and Novello, Tiago and Yukimura, Daniel and Magalhaes, Thales and Paz, Hallison and Velho, Luiz},
  booktitle={2021 34th SIBGRAPI Conference on Graphics, Patterns and Images (SIBGRAPI)},
  pages={17--24},
  year={2021},
  organization={IEEE}
}

@inproceedings{takikawa2021neural,
  author    = {Takikawa, Towaki and Litalien, Joey and Yin, Kangxue and Kreis, Karsten and Loop, Charles and Nowrouzezahrai, Derek and Jacobson, Alec and McGuire, Morgan and Fidler, Sanja},
  title     = {Neural Geometric Level of Detail: Real-Time Rendering With Implicit 3D Shapes},
  booktitle = {Proceedings of the IEEE/CVF Conference on Computer Vision and Pattern Recognition (CVPR)},
  month     = {June},
  year      = {2021},
  pages     = {11358-11367}
}

@inproceedings{kazhdan2006poisson,
  title={Poisson surface reconstruction},
  author={Kazhdan, Michael and Bolitho, Matthew and Hoppe, Hugues},
  booktitle={Proceedings of the fourth Eurographics symposium on Geometry processing},
  volume={7},
  year={2006}
}

@inproceedings{carr2001reconstruction,
  title={Reconstruction and representation of 3D objects with radial basis functions},
  author={Carr, Jonathan C and Beatson, Richard K and Cherrie, Jon B and Mitchell, Tim J and Fright, W Richard and McCallum, Bruce C and Evans, Tim R},
  booktitle={Proceedings of the 28th annual conference on Computer graphics and interactive techniques},
  pages={67--76},
  year={2001}
}

@article{paz2023mr,
  title={MR-Net: Multiresolution sinusoidal neural networks},
  author={Paz, Hallison and Perazzo, Daniel and Novello, Tiago and Schardong, Guilherme and Schirmer, Luiz and da Silva, Vinicius and Yukimura, Daniel and Chagas, Fabio and Lopes, Helio and Velho, Luiz},
  journal={Computers \& Graphics},
  year={2023},
  publisher={Elsevier}
}

@article{hart1996sphere,
  title={Sphere tracing: {A} geometric method for the antialiased ray tracing of implicit surfaces},
  author={Hart, John C},
  journal={The Visual Computer},
  volume={12},
  number={10},
  pages={527--545},
  year={1996},
  publisher={Citeseer}
}

@inproceedings{schoenberger2016sfm,
    author={Sch\"{o}nberger, Johannes Lutz and Frahm, Jan-Michael},
    title={Structure-from-Motion Revisited},
    booktitle={Conference on Computer Vision and Pattern Recognition (CVPR)},
    year={2016},
}

@article{knapitsch2017tanks,
  title={Tanks and temples: Benchmarking large-scale scene reconstruction},
  author={Knapitsch, Arno and Park, Jaesik and Zhou, Qian-Yi and Koltun, Vladlen},
  journal={ACM Transactions on Graphics (ToG)},
  volume={36},
  number={4},
  pages={1--13},
  year={2017},
  publisher={ACM New York, NY, USA}
}

@inproceedings{park2019deepsdf,
  title={Deepsdf: Learning continuous signed distance functions for shape representation},
  author={Park, Jeong Joon and Florence, Peter and Straub, Julian and Newcombe, Richard and Lovegrove, Steven},
  booktitle={Proceedings of the IEEE/CVF Conference on Computer Vision and Pattern Recognition},
  pages={165--174},
  year={2019}
}

@inproceedings{srinivasan2020lighthouse,
  title={Lighthouse: Predicting lighting volumes for spatially-coherent illumination},
  author={Srinivasan, Pratul P and Mildenhall, Ben and Tancik, Matthew and Barron, Jonathan T and Tucker, Richard and Snavely, Noah},
  booktitle={Proceedings of the IEEE/CVF Conference on Computer Vision and Pattern Recognition},
  pages={8080--8089},
  year={2020}
}

@article{zhang2021nerfactor,
  title={Nerfactor: Neural factorization of shape and reflectance under an unknown illumination},
  author={Zhang, Xiuming and Srinivasan, Pratul P and Deng, Boyang and Debevec, Paul and Freeman, William T and Barron, Jonathan T},
  journal={ACM Transactions on Graphics (ToG)},
  volume={40},
  number={6},
  pages={1--18},
  year={2021},
  publisher={ACM New York, NY, USA}
}

@inproceedings{dave2022pandora,
  title={Pandora: Polarization-aided neural decomposition of radiance},
  author={Dave, Akshat and Zhao, Yongyi and Veeraraghavan, Ashok},
  booktitle={European Conference on Computer Vision},
  pages={538--556},
  year={2022},
  organization={Springer}
}

@inproceedings{yariv2020idr,
 author = {Yariv, Lior and Kasten, Yoni and Moran, Dror and Galun, Meirav and Atzmon, Matan and Ronen, Basri and Lipman, Yaron},
 booktitle = {Advances in Neural Information Processing Systems},
 editor = {H. Larochelle and M. Ranzato and R. Hadsell and M.F. Balcan and H. Lin},
 pages = {2492--2502},
 publisher = {Curran Associates, Inc.},
 title = {Multiview Neural Surface Reconstruction by Disentangling Geometry and Appearance},
 volume = {33},
 year = {2020}
}

@article{zhang2020nerf++,
  title={Nerf++: Analyzing and improving neural radiance fields},
  author={Zhang, Kai and Riegler, Gernot and Snavely, Noah and Koltun, Vladlen},
  journal={arXiv preprint arXiv:2010.07492},
  year={2020}
}

@inproceedings{barron2021mip,
  title={Mip-nerf: A multiscale representation for anti-aliasing neural radiance fields},
  author={Barron, Jonathan T and Mildenhall, Ben and Tancik, Matthew and Hedman, Peter and Martin-Brualla, Ricardo and Srinivasan, Pratul P},
  booktitle={Proceedings of the IEEE/CVF international conference on computer vision},
  pages={5855--5864},
  year={2021}
}

@inproceedings{yu2021pixelnerf,
  title={pixelnerf: Neural radiance fields from one or few images},
  author={Yu, Alex and Ye, Vickie and Tancik, Matthew and Kanazawa, Angjoo},
  booktitle={Proceedings of the IEEE/CVF conference on computer vision and pattern recognition},
  pages={4578--4587},
  year={2021}
}

@inproceedings{zhang2021physg,
  title={Physg: Inverse rendering with spherical gaussians for physics-based material editing and relighting},
  author={Zhang, Kai and Luan, Fujun and Wang, Qianqian and Bala, Kavita and Snavely, Noah},
  booktitle={Proceedings of the IEEE/CVF Conference on Computer Vision and Pattern Recognition},
  pages={5453--5462},
  year={2021}
}

@inproceedings{mildenhall2020nerf,
  title={Nerf: Representing scenes as neural radiance fields for view synthesis},
  author={Mildenhall, Ben and Srinivasan, Pratul P and Tancik, Matthew and Barron, Jonathan T and Ramamoorthi, Ravi and Ng, Ren},
  booktitle={European conference on computer vision},
  pages={405--421},
  year={2020},
  organization={Springer}
}

@article{yang2021geometry,
  title={Geometry processing with neural fields},
  author={Yang, Guandao and Belongie, Serge and Hariharan, Bharath and Koltun, Vladlen},
  journal={Advances in Neural Information Processing Systems},
  volume={34},
  pages={22483--22497},
  year={2021}
}

@article{sundt2023marf,
  title={MARF: The medial atom ray field object representation},
  author={Sundt, Peder Bergebakken and Theoharis, Theoharis},
  journal={Computers \& Graphics},
  volume={115},
  pages={122--136},
  year={2023},
  publisher={Elsevier}
}

@inproceedings{lindell2022bacon,
  title={Bacon: Band-limited coordinate networks for multiscale scene representation},
  author={Lindell, David B and Van Veen, Dave and Park, Jeong Joon and Wetzstein, Gordon},
  booktitle={Proceedings of the IEEE/CVF conference on computer vision and pattern recognition},
  pages={16252--16262},
  year={2022}
}

@inproceedings{tiwary2023orca,
  title={ORCa: Glossy Objects as Radiance-Field Cameras},
  author={Tiwary, Kushagra and Dave, Akshat and Behari, Nikhil and Klinghoffer, Tzofi and Veeraraghavan, Ashok and Raskar, Ramesh},
  booktitle={Proceedings of the IEEE/CVF Conference on Computer Vision and Pattern Recognition},
  pages={20773--20782},
  year={2023}
}

@inproceedings{
jiao2023naisr,
title={\texttt{NAISR}: A 3D Neural Additive Model for Interpretable Shape Representation},
author={Yining Jiao and Carlton Jude ZDANSKI and Julia S Kimbell and Andrew Prince and Cameron P Worden and Samuel Kirse and Christopher Rutter and Benjamin Shields and William Alexander Dunn and Jisan Mahmud and Marc Niethammer},
booktitle={The Twelfth International Conference on Learning Representations},
year={2024},
url={https://openreview.net/forum?id=wg8NPfeMF9}
}

@article{silva2022mip,
  title={Mip-plicits: Level of detail factorization of neural implicits sphere tracing},
  author={da Silva, Vin{\'\i}cius and Novello, Tiago and Schardong, Guilherme and Schirmer, Luiz and Lopes, H{\'e}lio and Velho, Luiz},
  journal={arXiv preprint arXiv:2201.09147},
  volume={2},
  year={2022}
}

@article{yifan2021geometry,
  title={Geometry-consistent neural shape representation with implicit displacement fields},
  author={Yifan, Wang and Rahmann, Lukas and Sorkine-Hornung, Olga},
  journal={arXiv preprint arXiv:2106.05187},
  year={2021}
}

@inproceedings{chen2021learning,
  title={Learning continuous image representation with local implicit image function},
  author={Chen, Yinbo and Liu, Sifei and Wang, Xiaolong},
  booktitle={Proceedings of the IEEE/CVF conference on computer vision and pattern recognition},
  pages={8628--8638},
  year={2021}
}

@misc{schardong2024neural,
      title={Neural Implicit Morphing of Face Images},
      author={Guilherme Schardong and Tiago Novello and Hallison Paz and Iurii Medvedev and Vinícius da Silva and Luiz Velho and Nuno Gonçalves},
      year={2024},
      eprint={2308.13888},
      archivePrefix={arXiv},
      primaryClass={cs.CV}
}

@inproceedings{zheng2022imface,
  title={ImFace: A Nonlinear 3D Morphable Face Model with Implicit Neural Representations},
  author={Zheng, Mingwu and Yang, Hongyu and Huang, Di and Chen, Liming},
  booktitle={Proceedings of the IEEE/CVF Conference on Computer Vision and Pattern Recognition},
  pages={20343--20352},
  year={2022}
}

@article{zheng2023imfacepp,
  title={ImFace++: A Sophisticated Nonlinear 3D Morphable Face Model with Implicit Neural Representations},
  author={Zheng, Mingwu and Zhang, Haiyu and Yang, Hongyu and Chen, Liming and Huang, Di},
  journal={arXiv preprint arXiv:2312.04028},
  year={2023}
}

@InProceedings{Schirmer2023how,
               author = {Schirmer, Luiz and Novello, Tiago and Schardong, Guilherme and
                         Silva, Vin{\'{\i}}cius da and Lopes, H{\'e}lio and Velho,
                         Luiz},
                title = {How to train your (neural) dragon},
            booktitle = {Conference on Graphics, Patterns and Images},
                 year = {2023},
                  url = {http://urlib.net/ibi/8JMKD3MGPEW34M/49T46US}
}

@inproceedings{novello2023neural,
  title={Neural Implicit Surface Evolution},
  author={Novello, Tiago and da Silva, Vinicius and Schardong, Guilherme and Schirmer, Luiz and Lopes, Helio and Velho, Luiz},
  booktitle={Proceedings of the IEEE/CVF International Conference on Computer Vision},
  pages={14279--14289},
  year={2023}
}

@inproceedings{paz2022,
  title={Multiresolution Neural Networks for Imaging},
  author = {Paz, Hallison and Novello, Tiago and Silva, Vinicius and Schirmer, Luiz and Schardong, Guilherme and Chagas, Fabio and Lopes, Helio and Velho, Luiz},
  booktitle={Proceedings of 35th Conference on Graphics, Patterns and Images (SIBGRAPI)},
  note={to appear},
  year={2022}
}

@inproceedings{saragadam2022miner,
  title={Miner: Multiscale implicit neural representation},
  author={Saragadam, Vishwanath and Tan, Jasper and Balakrishnan, Guha and Baraniuk, Richard G and Veeraraghavan, Ashok},
  booktitle={European Conference on Computer Vision},
  pages={318--333},
  year={2022},
  organization={Springer}
}

@inproceedings{marschner2023constructive,
  title={Constructive Solid Geometry on Neural Signed Distance Fields},
  author={Marschner, Zo{\"e} and Sell{\'a}n, Silvia and Liu, Hsueh-Ti Derek and Jacobson, Alec},
  booktitle={SIGGRAPH Asia 2023 Conference Papers},
  pages={1--12},
  year={2023}
}

@article{perazzo2023directvoxgo++,
  title={DirectVoxGO++: Grid-based fast object reconstruction using radiance fields},
  author={Perazzo, Daniel and Lima, Jo{\~a}o Paulo and Velho, Luiz and Teichrieb, Veronica},
  journal={Computers \& Graphics},
  volume={114},
  pages={96--104},
  year={2023},
  publisher={Elsevier}
}

@article{paz2024implicit,
  title={Implicit Neural Representation of Tileable Material Textures},
  author={Paz, Hallison and Novello, Tiago and Velho, Luiz},
  journal={arXiv preprint arXiv:2402.02208},
  year={2024}
}

@inproceedings{mescheder2019occupancy,
  title={Occupancy networks: Learning 3d reconstruction in function space},
  author={Mescheder, Lars and Oechsle, Michael and Niemeyer, Michael and Nowozin, Sebastian and Geiger, Andreas},
  booktitle={Proceedings of the IEEE/CVF conference on computer vision and pattern recognition},
  pages={4460--4470},
  year={2019}
}

@inproceedings{Oechsle2021ICCV,
            title = {UNISURF: Unifying Neural Implicit Surfaces and Radiance Fields for Multi-View Reconstruction},
            author = {Oechsle, Michael and Peng, Songyou and Geiger, Andreas},
            booktitle = {International Conference on Computer Vision (ICCV)},
            year = {2021},
            doi = {}
 }

@inproceedings{wang2023neus2,
  title={Neus2: Fast learning of neural implicit surfaces for multi-view reconstruction},
  author={Wang, Yiming and Han, Qin and Habermann, Marc and Daniilidis, Kostas and Theobalt, Christian and Liu, Lingjie},
  booktitle={Proceedings of the IEEE/CVF International Conference on Computer Vision},
  pages={3295--3306},
  year={2023}
}

@article{kajiya1984ray,
  title={Ray tracing volume densities},
  author={Kajiya, James T and Von Herzen, Brian P},
  journal={ACM SIGGRAPH computer graphics},
  volume={18},
  number={3},
  pages={165--174},
  year={1984},
  publisher={ACM New York, NY, USA}
}

@article{max1995optical,
  title={Optical models for direct volume rendering},
  author={Max, Nelson},
  journal={IEEE Transactions on Visualization and Computer Graphics},
  volume={1},
  number={2},
  pages={99--108},
  year={1995},
  publisher={IEEE}
}

@inproceedings{li2023neuralangelo,
  title={Neuralangelo: High-fidelity neural surface reconstruction},
  author={Li, Zhaoshuo and M{\"u}ller, Thomas and Evans, Alex and Taylor, Russell H and Unberath, Mathias and Liu, Ming-Yu and Lin, Chen-Hsuan},
  booktitle={Proceedings of the IEEE/CVF Conference on Computer Vision and Pattern Recognition},
  pages={8456--8465},
  year={2023}
}

@article{yu2022monosdf,
  title={Monosdf: Exploring monocular geometric cues for neural implicit surface reconstruction},
  author={Yu, Zehao and Peng, Songyou and Niemeyer, Michael and Sattler, Torsten and Geiger, Andreas},
  journal={Advances in neural information processing systems},
  volume={35},
  pages={25018--25032},
  year={2022}
}

@inproceedings{yariv2023bakedsdf,
  title={Bakedsdf: Meshing neural sdfs for real-time view synthesis},
  author={Yariv, Lior and Hedman, Peter and Reiser, Christian and Verbin, Dor and Srinivasan, Pratul P and Szeliski, Richard and Barron, Jonathan T and Mildenhall, Ben},
  booktitle={ACM SIGGRAPH 2023 Conference Proceedings},
  pages={1--9},
  year={2023}
}

@inproceedings{wang2021neus,
 author = {Wang, Peng and Liu, Lingjie and Liu, Yuan and Theobalt, Christian and Komura, Taku and Wang, Wenping},
 booktitle = {Advances in Neural Information Processing Systems},
 editor = {M. Ranzato and A. Beygelzimer and Y. Dauphin and P.S. Liang and J. Wortman Vaughan},
 pages = {27171--27183},
 publisher = {Curran Associates, Inc.},
 title = {NeuS: Learning Neural Implicit Surfaces by Volume Rendering for Multi-view Reconstruction},
 volume = {34},
 year = {2021}
}

@inproceedings{yariv2021volume,
 author = {Yariv, Lior and Gu, Jiatao and Kasten, Yoni and Lipman, Yaron},
 booktitle = {Advances in Neural Information Processing Systems},
 editor = {M. Ranzato and A. Beygelzimer and Y. Dauphin and P.S. Liang and J. Wortman Vaughan},
 pages = {4805--4815},
 publisher = {Curran Associates, Inc.},
 title = {Volume Rendering of Neural Implicit Surfaces},
 volume = {34},
 year = {2021}
}

@inproceedings{long2022sparseneus,
  title={Sparseneus: Fast generalizable neural surface reconstruction from sparse views},
  author={Long, Xiaoxiao and Lin, Cheng and Wang, Peng and Komura, Taku and Wang, Wenping},
  booktitle={European Conference on Computer Vision},
  pages={210--227},
  year={2022},
  organization={Springer}
}

@inproceedings{darmon2022improving,
  title={Improving neural implicit surfaces geometry with patch warping},
  author={Darmon, Fran{\c{c}}ois and Bascle, B{\'e}n{\'e}dicte and Devaux, Jean-Cl{\'e}ment and Monasse, Pascal and Aubry, Mathieu},
  booktitle={Proceedings of the IEEE/CVF Conference on Computer Vision and Pattern Recognition},
  pages={6260--6269},
  year={2022}
}

@inproceedings{ma2023towards,
  title={Towards better gradient consistency for neural signed distance functions via level set alignment},
  author={Ma, Baorui and Zhou, Junsheng and Liu, Yu-Shen and Han, Zhizhong},
  booktitle={Proceedings of the IEEE/CVF Conference on Computer Vision and Pattern Recognition},
  pages={17724--17734},
  year={2023}
}

@article{muller2022instant,
  title={Instant neural graphics primitives with a multiresolution hash encoding},
  author={M{\"u}ller, Thomas and Evans, Alex and Schied, Christoph and Keller, Alexander},
  journal={ACM transactions on graphics (TOG)},
  volume={41},
  number={4},
  pages={1--15},
  year={2022},
  publisher={ACM New York, NY, USA}
}

@inproceedings{li2023nerfacc,
  title={Nerfacc: Efficient sampling accelerates nerfs},
  author={Li, Ruilong and Gao, Hang and Tancik, Matthew and Kanazawa, Angjoo},
  booktitle={Proceedings of the IEEE/CVF International Conference on Computer Vision},
  pages={18537--18546},
  year={2023}
}

@article{zhao2022human,
  title={Human performance modeling and rendering via neural animated mesh},
  author={Zhao, Fuqiang and Jiang, Yuheng and Yao, Kaixin and Zhang, Jiakai and Wang, Liao and Dai, Haizhao and Zhong, Yuhui and Zhang, Yingliang and Wu, Minye and Xu, Lan and others},
  journal={ACM Transactions on Graphics (TOG)},
  volume={41},
  number={6},
  pages={1--17},
  year={2022},
  publisher={ACM New York, NY, USA}
}

@inproceedings{jensen2014large,
  title={Large scale multi-view stereopsis evaluation},
  author={Jensen, Rasmus and Dahl, Anders and Vogiatzis, George and Tola, Engin and Aan{\ae}s, Henrik},
  booktitle={Proceedings of the IEEE conference on computer vision and pattern recognition},
  pages={406--413},
  year={2014}
}

@inproceedings{fainstein2024dudf,
  title={DUDF: Differentiable Unsigned Distance Fields with Hyperbolic Scaling},
  author={Fainstein, Miguel and Siless, Viviana and Iarussi, Emmanuel},
  booktitle={Proceedings of the IEEE/CVF Conference on Computer Vision and Pattern Recognition},
  pages={4484--4493},
  year={2024}
}

@article{ullman1979interpretation,
  title={The interpretation of structure from motion},
  author={Ullman, Shimon},
  journal={Proceedings of the Royal Society of London. Series B. Biological Sciences},
  volume={203},
  number={1153},
  pages={405--426},
  year={1979},
  publisher={The Royal Society London}
}

@inproceedings{lowe1999object,
  title={Object recognition from local scale-invariant features},
  author={Lowe, David G},
  booktitle={Proceedings of the seventh IEEE international conference on computer vision},
  volume={2},
  pages={1150--1157},
  year={1999},
  organization={Ieee}
}

@article{fischler1981random,
  title={Random sample consensus: a paradigm for model fitting with applications to image analysis and automated cartography},
  author={Fischler, Martin A and Bolles, Robert C},
  journal={Communications of the ACM},
  volume={24},
  number={6},
  pages={381--395},
  year={1981},
  publisher={ACM New York, NY, USA}
}

@inproceedings{mederos2003moving,
  title={Moving least squares multiresolution surface approximation},
  author={Mederos, Boris and Velho, Luiz and De Figueiredo, Luiz Henrique},
  booktitle={16th Brazilian symposium on computer graphics and image processing (SIBGRAPI 2003)},
  pages={19--26},
  year={2003},
  organization={IEEE}
}

@inproceedings{williams1983pyramidal,
  title={Pyramidal parametrics},
  author={Williams, Lance},
  booktitle={Proceedings of the 10th annual conference on Computer graphics and interactive techniques},
  pages={1--11},
  year={1983}
}

@article{mallat1989theory,
  title={A theory for multiresolution signal decomposition: the wavelet representation},
  author={Mallat, Stephane G},
  journal={IEEE transactions on pattern analysis and machine intelligence},
  volume={11},
  number={7},
  pages={674--693},
  year={1989},
  publisher={Ieee}
}

@inproceedings{shabanov2024banf,
  title={BANF: Band-limited Neural Fields for Levels of Detail Reconstruction},
  author={Shabanov, Akhmedkhan and Govindarajan, Shrisudhan and Reading, Cody and Goli, Lily and Rebain, Daniel and Yi, Kwang Moo and Tagliasacchi, Andrea},
  booktitle={Proceedings of the IEEE/CVF Conference on Computer Vision and Pattern Recognition},
  pages={20571--20580},
  year={2024}
}

@misc{Yu2022SDFStudio,
    author    = {Yu, Zehao and Chen, Anpei and Antic, Bozidar and Peng, Songyou and Bhattacharyya, Apratim
                 and Niemeyer, Michael and Tang, Siyu and Sattler, Torsten and Geiger, Andreas},
    title     = {SDFStudio: A Unified Framework for Surface Reconstruction},
    year      = {2022},
    url       = {https://github.com/autonomousvision/sdfstudio},
}

@inproceedings{nerfstudio,
    title        = {Nerfstudio: A Modular Framework for Neural Radiance Field Development},
    author       = {
        Tancik, Matthew and Weber, Ethan and Ng, Evonne and Li, Ruilong and Yi, Brent
        and Kerr, Justin and Wang, Terrance and Kristoffersen, Alexander and Austin,
        Jake and Salahi, Kamyar and Ahuja, Abhik and McAllister, David and Kanazawa,
        Angjoo
    },
    year         = 2023,
    booktitle    = {ACM SIGGRAPH 2023 Conference Proceedings},
    series       = {SIGGRAPH '23}
}

@article{fu2022geo,
  title={Geo-neus: Geometry-consistent neural implicit surfaces learning for multi-view reconstruction},
  author={Fu, Qiancheng and Xu, Qingshan and Ong, Yew Soon and Tao, Wenbing},
  journal={Advances in Neural Information Processing Systems},
  volume={35},
  pages={3403--3416},
  year={2022}
}

@incollection{lorensen1998marching,
  title={Marching cubes: A high resolution 3D surface construction algorithm},
  author={Lorensen, William E and Cline, Harvey E},
  booktitle={Seminal graphics: pioneering efforts that shaped the field},
  pages={347--353},
  year={1998},
  publisher={ACM}
}

@inproceedings{azinovic2022neural,
  title={Neural rgb-d surface reconstruction},
  author={Azinovi{\'c}, Dejan and Martin-Brualla, Ricardo and Goldman, Dan B and Nie{\ss}ner, Matthias and Thies, Justus},
  booktitle={Proceedings of the IEEE/CVF Conference on Computer Vision and Pattern Recognition},
  pages={6290--6301},
  year={2022}
}
\end{document}